\documentclass[a4paper,fleqn]{cas-dc}

\usepackage[numbers]{natbib}
\usepackage{graphicx}%

\usepackage{amsmath,amssymb,amsfonts}%
\usepackage{amsthm}%
\usepackage{mathrsfs}%
\usepackage[title]{appendix}%
\usepackage{xcolor}%
\usepackage{textcomp}%
\usepackage{manyfoot}%
\usepackage{booktabs}%
\usepackage{algorithm}%
\usepackage{algorithmicx}%
\usepackage{algpseudocode}%
\usepackage{listings}%
\usepackage{multirow}%
\usepackage{subcaption}

\begin{document}
\let\WriteBookmarks\relax
\def\floatpagepagefraction{1}
\def\textpagefraction{.001}
\shorttitle{ShadowMamba: State-Space Model for Shadow Removal}
\shortauthors{Zhu Xiujin et~al.}

\title [mode = title]{ShadowMamba: State-Space Model with Boundary-Region Selective Scan for Shadow Removal}

\author[1]{Zhu Xiujin}[type=editor,
                       orcid=0009-0002-1017-2474]
\credit{Conceptualization, Methodology, Writing - Original draft preparation}
\ead{s2121087@siswa.um.edu.my}

\affiliation[1]{organization={Department of Electrical Engineering, Faculty of Engineering, Universiti Malaya},
                addressline={Lembah Pantai}, 
                city={Kuala Lumpur},
                postcode={50603}, 
                country={Malaysia}}

\affiliation[2]{organization={Faculty of Engineering and Information Technology, Southern University College},
            city={Skudai},
            postcode={81300}, 
            country={Malaysia}}

\author[1]{Chow Chee-Onn}[orcid=0000-0001-6044-2650]
\cormark[1]
\ead{cochow@um.edu.my}
\cortext[cor1]{Corresponding author}
\credit{Supervision, Review, Funding acquisition}

\author[1,2]{Chuah Joon Huang}[orcid=0000-0001-9058-3497]
\ead{jhchuah@um.edu.my}
\credit{Supervision}

\begin{abstract}
Image shadow removal is a typical low-level vision task. Shadows cause local brightness shifts, which reduce the performance of downstream vision tasks. Currently, Transformer-based shadow removal methods suffer from quadratic computational complexity due to the self-attention mechanism. To improve efficiency, many approaches use local attention, but this limits the ability to model global information and weakens the perception of brightness changes between regions. Recently, Mamba has shown strong performance in vision tasks by enabling global modeling with linear complexity. However, existing scanning strategies are not suitable for shadow removal, as they ignore the semantic continuity of shadow boundaries and internal regions. To address this, this paper proposes a boundary-region selective scanning mechanism that captures local details while enhancing semantic continuity between them, effectively improving shadow removal performance. In addition, a shadow mask denoising method is introduced to support the scanning mechanism and improve data quality. Based on these techniques, this paper presents a model called ShadowMamba, the first Mamba-based model designed for shadow removal. Experimental results show that the proposed method outperforms existing mainstream approaches on the AISTD, ISTD, and SRD datasets, and also offers clear advantages in parameter efficiency and computational complexity. Code is available at: \url{https://github.com/ZHUXIUJINChris/ShadowMamba}.
\end{abstract}


\maketitle

\section{Introduction}
\label{sec1}
Shadows are cast when objects block light, making it inevitable that they are captured in images during acquisition. The presence of shadows not only causes the image to lose certain information but also affects the accuracy of downstream tasks, such as object detection \cite{zhao2024detrs, chen2023diffusiondet}, instance segmentation \cite{jain2023oneformer}, and image classification \cite{touvron2022resmlp}. Consequently, effective shadow removal is essential for enhancing image quality and improving the performance of these vision tasks.

Traditional shadow removal methods are mainly categorized into illumination transfer methods \cite{guo2012paired, zhang2015shadow} and shadow region relighting methods \cite{gong2014interactive}. These approaches rely on the physical modeling of shadows and are effective for handling single shadow types. However, they often struggle in complex background scenarios. In recent years, deep learning progressively replaces traditional techniques, giving rise to numerous shadow removal methods \cite{le2021physics, li2023leveraging, guo2023shadowformer, guo2023shadowdiffusion, liu2024recasting, xiao2024homoformer} based on CNN and Transformer architectures. These methods model shadow images by leveraging brightness information \cite{zhu2022efficient, einy2022physics}, shadow boundary information \cite{niu2022boundary, guo2023boundary}, or regional characteristics \cite{chen2021canet, wan2024crformer, guo2023shadowformer}, employing deep networks to achieve effective shadow removal.

Currently, Transformer-based methods are widely used in shadow removal tasks, mainly because the self-attention mechanism can flexibly model pixel relationships between regions, enabling illumination transfer. However, the dot-product operations in self-attention have quadratic computational complexity, which limits the efficiency of ViT-based models like CRFormer \cite{wan2024crformer} when processing high-resolution images. To reduce complexity, ShadowFormer \cite{guo2023shadowformer} and HomoFormer \cite{xiao2024homoformer} introduce local attention mechanisms \cite{liu2021swin,  xiao2023random}. Nevertheless, these mechanisms can only compute self-attention over a portion of the tokens at a time, requiring a trade-off between receptive field size and computational efficiency. This limits the model’s ability to capture long-range dependencies and leads to suboptimal performance.

Recently, an improved structured state-space sequence model called Mamba \cite{gu2023mamba} was proposed. This model has built long-range dependencies with linear complexity, has a global receptive field, and introduces a new form of attention through a selective scanning mechanism. Mamba was originally designed for one-dimensional sequence modeling, while visual tasks need to consider pixel relationships in the upward, downward, left, and right directions to maintain spatial continuity. Many Mamba-based methods \cite{hu2024zigma, huang2024localmamba, shi2025vmambair} have significantly improved Mamba's performance on image tasks by optimizing scanning paths and enhancing semantic relevance.

Shadow removal mainly relies on information from non-shadow regions to adjust the brightness in shadowed areas. Therefore, the model needs a large receptive field to effectively capture brightness differences across regions. Although current Mamba-based scanning mechanisms \cite{zhu2024vision, Liu2024VMambaVS, guo2024mambair, shi2025vmambair} have global modeling capability, their attention to local details remains limited. In shadow removal tasks, handling local details and boundaries is especially important, as improper processing can easily lead to artifacts and restoration errors. To enhance the model’s ability to capture local information, introducing a local scanning mechanism is necessary. LocalMamba \cite{huang2024localmamba} uses a window-based approach to strengthen the semantic continuity of local pixels and scans all windows in sequence. However, this strategy partly breaks the semantic continuity at shadow boundaries. In addition, pixels with similar brightness usually have strong semantic relevance, meaning pixels within the same region are more closely related. This scanning method may also disrupt such internal associations.

\begin{figure}[!t]
\centering
\subfloat[Local scan \cite{huang2024localmamba}]{
		\includegraphics[width=0.22\textwidth,height=0.14\textheight]{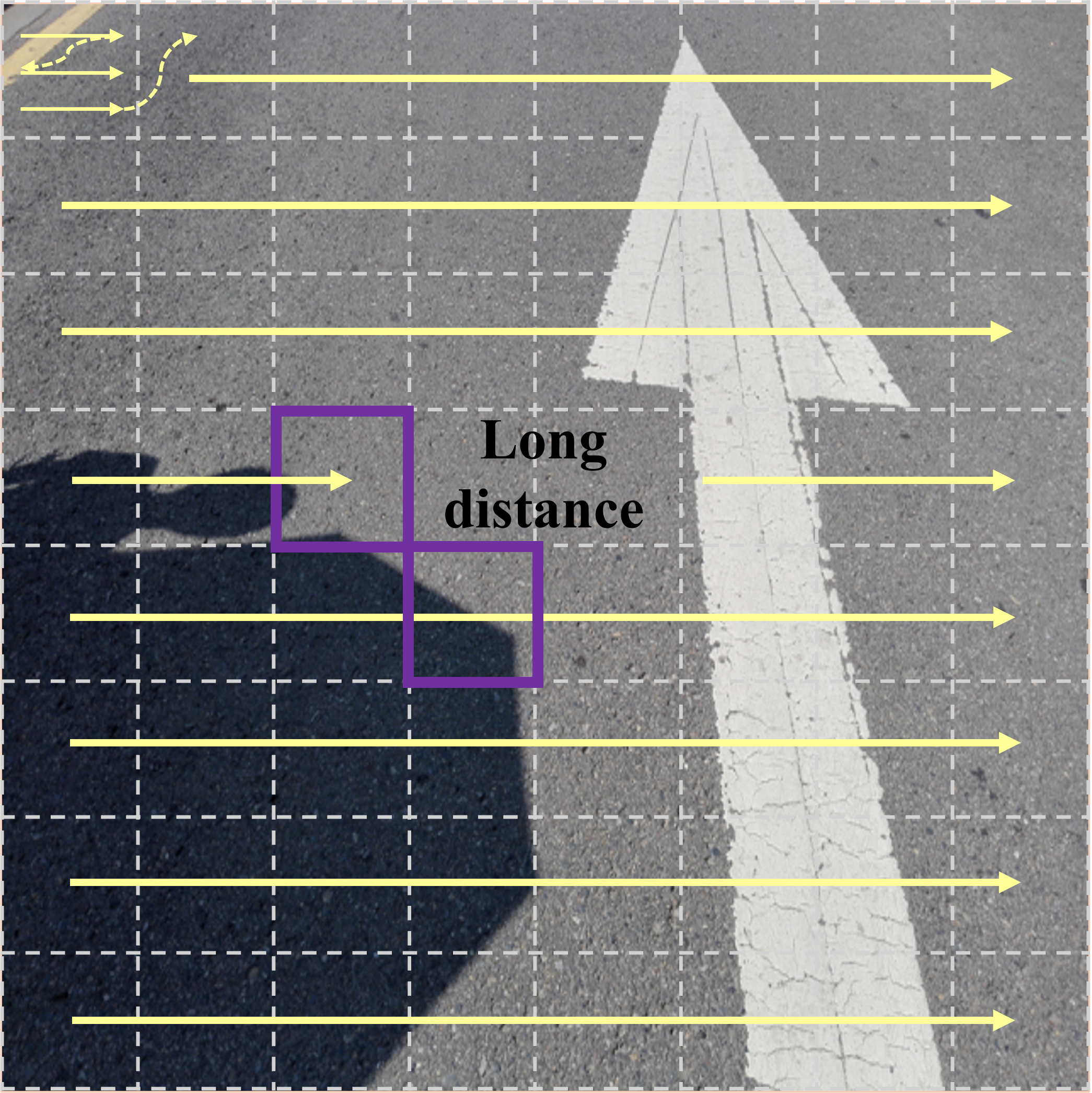}
\label{Fig 1(a)}}
\hspace{0.15cm}
\subfloat[\textbf{Boundary-region scan}]{
		\includegraphics[width=0.22\textwidth,height=0.14\textheight]{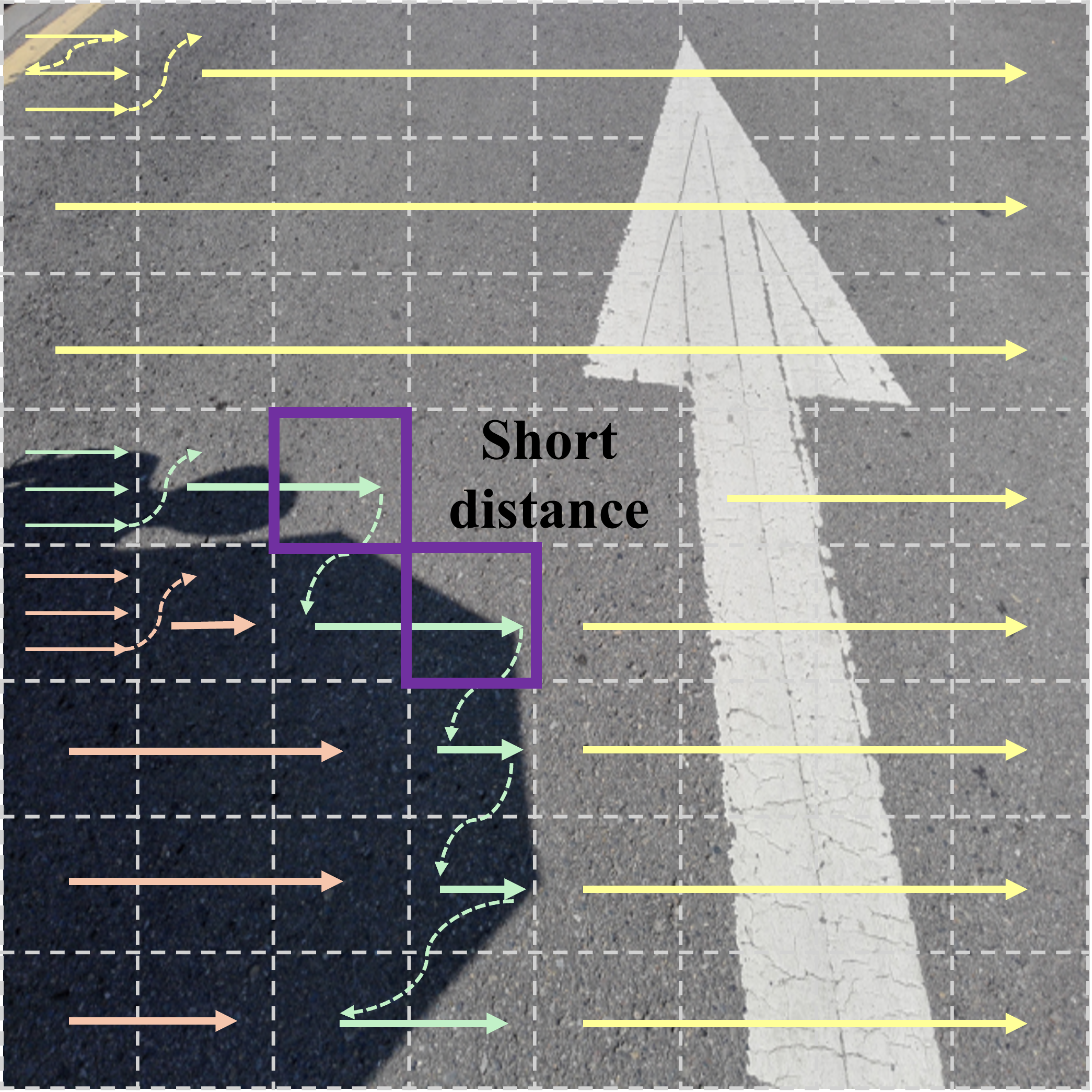}
\label{Fig 1(b)}}
\caption{The two purple windows belong to the boundary region of the shadow and exhibit a strong semantic correlation, so their distance in the long sequence should be as close as possible. Regardless of whether the scanning is horizontal or vertical, the proposed boundary-region scanning mechanism effectively reduces their distance in the long sequence. (This is only a rough demonstration, and the actual windows are much smaller.)}
\label{Fig 1}
\end{figure}

To address the above issue, this paper proposes a novel boundary-region selective scanning mechanism. This method introduces an additional mask input to divide the shadow image into multiple windows, following a local scanning strategy \cite{huang2024localmamba}. According to a set of predefined rules, these windows are classified into shadow regions, boundary regions, and non-shadow regions. These three types of regions are then scanned in order, which effectively enhances the semantic continuity among pixels of the same type. By placing pixels of the same category closer together in the long sequence, the modeling capability of the state-space model is improved. Figure \ref{Fig 1} provides a detailed comparison between the boundary-region scanning mechanism and the local scanning mechanism.

In addition, this paper also designs a shadow mask denoising method specifically for the boundary-region selective scanning mechanism. This method effectively removes noise from non-shadow regions in the mask without affecting the shadow boundaries, ensuring the effectiveness of the scanning mechanism and improving mask accuracy. As a result, the performance of shadow removal methods that rely on mask guidance is enhanced. Finally, based on the boundary-region selective scanning mechanism and the global cross-scanning mechanism, the Boundary Region State-Space Block (BRSSB) and the Global State-Space Block (GSSB) to extract local detail information and global brightness information, respectively. These two blocks are integrated through a hierarchical U-Net architecture to build ShadowMamba, the first Mamba-based model specifically designed for shadow removal tasks. The main contributions of this work are summarized as follows: 

\begin{itemize}
\item 
This paper is the first to apply the state-space sequence model Mamba to the shadow removal task by designing ShadowMamba, which achieves global token modeling with linear complexity and effectively addresses both hard and soft shadows.

\item 
This paper designs a boundary-region selective scanning mechanism to capture local detail information and improves the semantic continuity among local details in the same region, which helps the model better understand local features.

\item 
This paper proposes a shadow mask denoising method, which allows the scanning mechanism to work effectively even when the mask contains noise. The method also improves the performance of other models that rely on mask guidance.

\item
ShadowMamba outperforms existing state-of-the-art (SOTA) methods on standard datasets such as AISTD, ISTD, and SRD, while maintaining lower parameter count and computational complexity.
\end{itemize}

\section{Related work}
\label{sec2}

\subsection{Image shadow removal}

There are various CNN-based methods for shadow removal. EMDN \cite{zhu2022efficient} and PBID \cite{einy2022physics} use a linear illumination model to estimate the illumination parameters of shadow regions. SG-ShadowNet \cite{wan2022style} uses a style-guided mechanism to transfer feature information from non-shadow regions to shadow regions. BM-Net \cite{zhu2022bijective} uses shadow generation to assist the shadow removal process, guiding it with invariant image information. BA-ShadowNet \cite{niu2022boundary} leverages boundary information for shadow removal, demonstrating the effectiveness of boundary information in shadow removal. Inpaint4shadow \cite{li2023leveraging} models the shadow removal task as a combination of image inpainting and shadow elimination. DMTN \cite{liu2023decoupled} performs shadow removal by explicitly decoupling features. StructNet \cite{liu2023structure} adopts a two-stage approach to perform shadow removal at both the structural and pixel levels.

Many shadow removal methods are built on Transformer architectures. CRFormer \cite{wan2024crformer} and ShadowFormer \cite{guo2023shadowformer} reweight the attention matrix using shadow masks to enhance information interaction between different regions. However, this approach relies heavily on the accuracy of the masks. HomoFormer \cite{xiao2024homoformer} adopts a random pixel shuffle strategy to enable global information interaction within each window, but it also introduces some unnecessary redundant computation. OmniSR \cite{xu2025omnisr} integrates semantic features and geometric information, reweighting features within the local attention mechanism to achieve more precise shadow removal. ShadowMaskFormer \cite{li2025shadowmaskformer} introduces shadow mask information during the patch embedding stage, enhancing the model's focus on shadow regions.

In addition, generative models based on the aforementioned architectures are emerging, with Diffusion-based shadow removal methods gradually replacing GAN-based approaches. ShadowDiffusion \cite{guo2023shadowdiffusion} leverages generative and degradation priors to progressively predict more accurate shadow masks, assisting in shadow removal and achieving outstanding results. LFG-Diffusion \cite{mei2024latent} minimizes the difference between the feature spaces of shadow and shadow-free images to obtain latent features, which contain more useful information than explicit features. DeS3 \cite{jin2024des3} excels at handling self-shadows by introducing ViT similarity loss, which enables more robust extraction of structural information. RRLSR \cite{liu2024recasting} uses illumination decomposition and a bidirectional correction network to achieve high-quality shadow removal.

\subsection{State space models}

Recently, State-Space Models (SSMs) have gained significant attention in the field of deep learning, emerging as a strong competitor to Transformers. SSMs originate from control theory and are used to describe the state transitions of sequences. To adapt them to deep learning tasks, Hippo \cite{gu2020hippo} discretizes the original model and enables parallel computation, while S4 \cite{gu2021efficiently} further improves computational efficiency. Building on this, Mamba \cite{gu2023mamba} combines a selective scanning mechanism with hardware-aware optimization, enabling linear-complexity global modeling and outperforming Transformers on most tasks.

The introduction of Mamba has promoted its wide application in tasks such as natural language processing \cite{tao2024scaling}, speech recognition \cite{zhang2025mamba}, and image and video processing. However, since its original scanning mechanism is designed for one-dimensional (1D) data, it struggles to effectively model spatial continuity in visual tasks. As a result, many studies focus on improving its scanning strategy to extend its applicability to two-dimensional (2D) and three-dimensional (3D) data. For example, Vim \cite{zhu2024vision} and VMamba \cite{Liu2024VMambaVS} introduce bidirectional and cross-scanning mechanisms, while ZigMa \cite{hu2024zigma} uses zigzag scanning to enhance spatial structure awareness. LocalMamba \cite{huang2024localmamba} strengthens local detail modeling through a local window scanning strategy. Videomamba \cite{li2024videomamba} and Vivim \cite{yang2024vivim} further introduce 3D spatiotemporal scanning mechanisms to improve Mamba's performance in video tasks.

In the field of image restoration, many tasks have adopted the Mamba architecture and achieved promising results. MambaIR \cite{guo2024mambair} improves restoration performance by introducing local enhancement and channel attention mechanisms. VmambaIR \cite{shi2025vmambair} proposes omnidirectional selective scanning to model information flow across multiple feature dimensions. FreqMamba \cite{zou2024freqmamba} performs local scanning in the frequency domain and integrates Fourier transform to generate rain-free images. RainMamba \cite{wu2024rainmamba} introduces a Hilbert scanning mechanism that more accurately captures sequence-level local information in videos. These studies show that optimizing the scanning mechanism can also effectively improve image restoration performance.

\section{Methodology}

\subsection{Architecture}
\begin{figure*}[t!]

\centering
\includegraphics[width=0.85\textwidth,height=0.25\textheight]{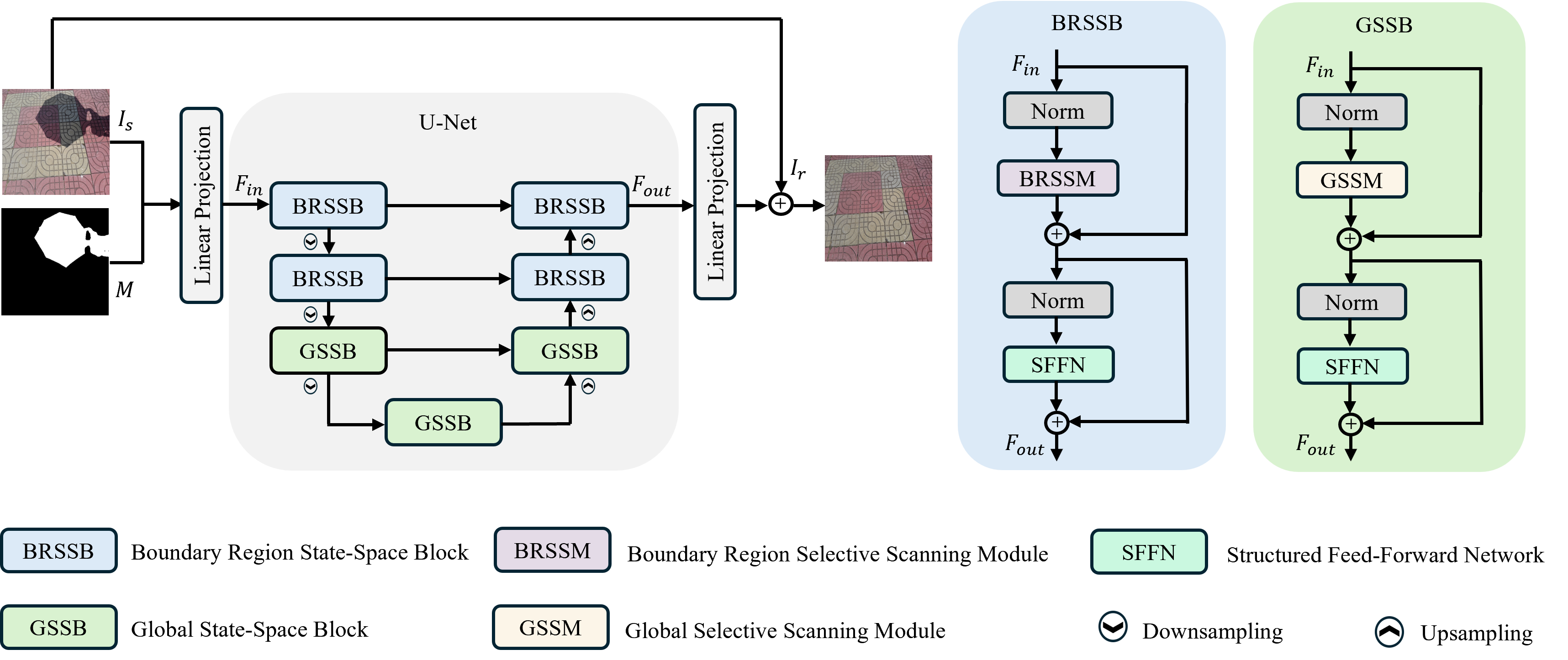}
\vspace{1mm}
\caption{The architecture of the proposed ShadowMamba.}
\label{Fig 2}

\end{figure*}

The overall pipeline of the proposed ShadowMamba is shown in Figure \ref{Fig 2}. This method builds a shadow removal network based on the Mamba architecture and adopts a U-Net structure to capture shadow features at different scales. The network consists of multiple Boundary Region State-Space Blocks (BRSSBs) and Global State-Space Blocks (GSSBs), which are used to extract local detail information and global brightness features from shadow images, respectively. It is well known that the shallow layers of the U-Net architecture focus on extracting local details of the image, while BRSSB excels at capturing local information and enhances the model's understanding of details through the boundary-region scanning mechanism. Therefore, BRSSB is placed in the first two layers of the U-Net. In the deep layers of the U-Net, as the spatial resolution of the feature maps decreases, the model is more suitable for capturing overall brightness features from a global perspective. With this hierarchical structure, ShadowMamba effectively reduces the number of parameters and computational complexity.

Specifically, given a shadow input $I_s\in R^{3\times H \times W}$ and its corresponding shadow mask $M\in R^{1\times H \times W}$, an overlapped embedding  is first applied to obtain the shallow feature $F_{in}\in R^{C\times H \times W}$ from the input. Subsequently, these shallow features are fed into a U-Net architecture composed of multiple BRSSBs and GSSBs, with the decoded output being $F_{out}\in R^{C\times H \times W}$. Finally, a linear mapping layer is applied to obtain the residual $I_r\in R^{3\times H \times W}$ between the output image and the input image.

\subsubsection{\textbf{Boundary region state-space block (BRSSB)}}

BRSSB follows the classic Transformer architecture design, as shown in the blue section of Figure \ref{Fig 2}. After the first normalization layer, it integrates the Boundary Region Selective Scaning Module (BRSSM), which incorporates a boundary-region selective scanning mechanism. As the core component of BRSSB, this module effectively captures boundary information and local details within shadow images. After the second normalization layer, the Structured Feed-Forward Network (SFFN) is introduced, primarily to perform structured modeling on the disrupted sequences, enhancing the model’s representation capability.

\begin{figure*}[!t]
\centering

\subfloat[]{
		\includegraphics[width=0.45\textwidth,height=0.19\textheight]{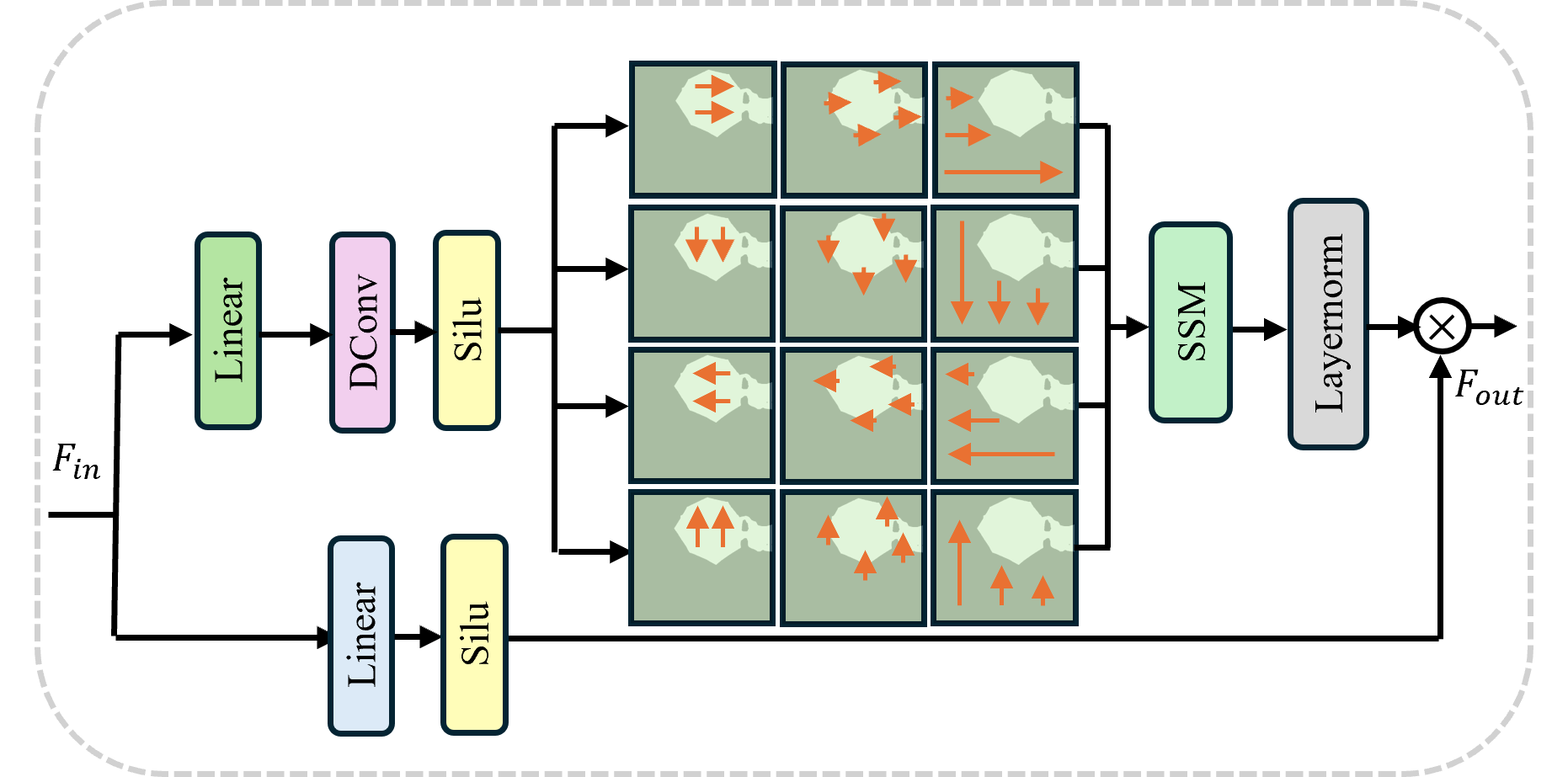}
\label{Fig 3a}}
\subfloat[]{
		\includegraphics[width=0.35\textwidth,height=0.19\textheight]{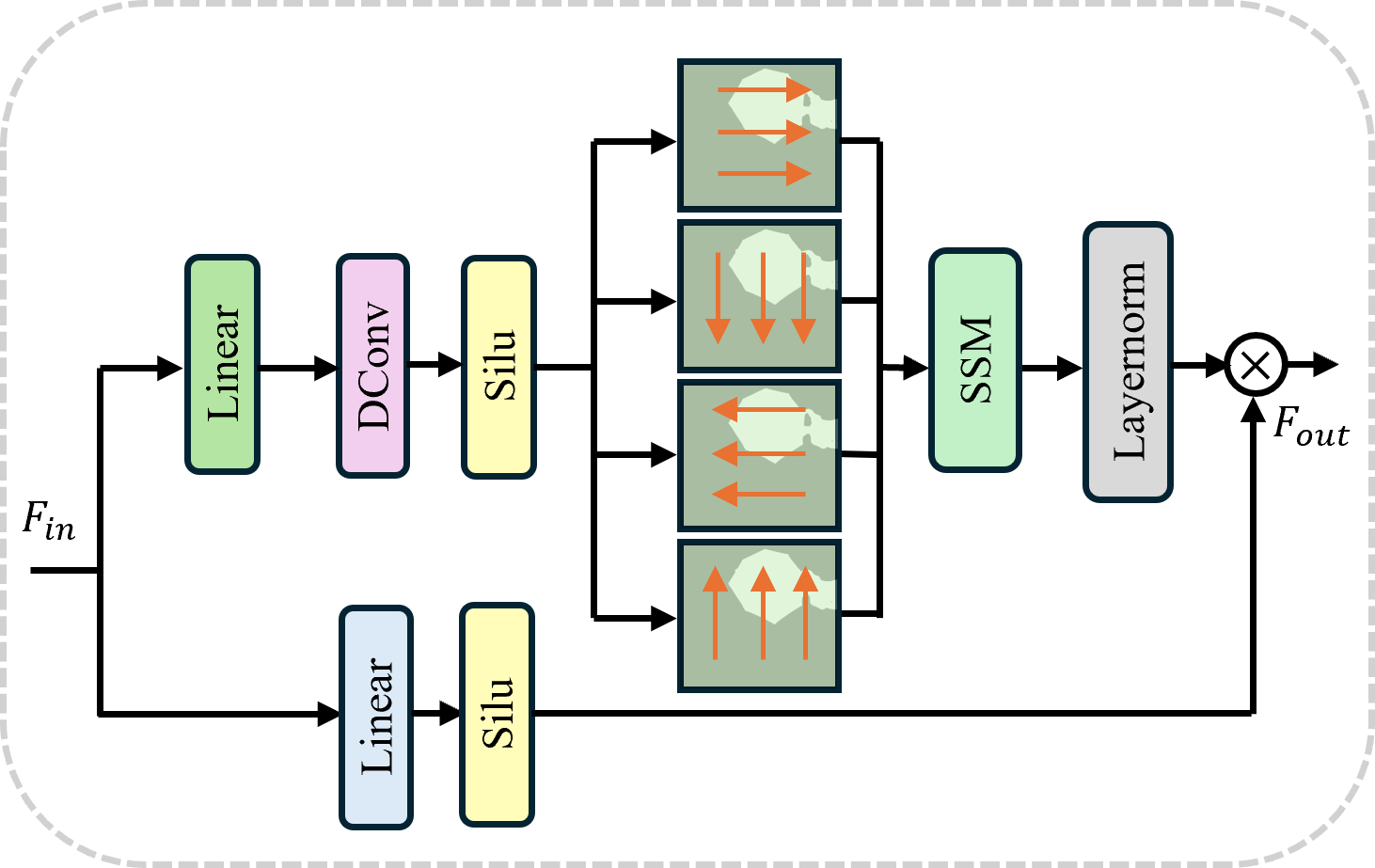}
\label{Fig 3b}}
\subfloat[]{
		\includegraphics[width=0.15\textwidth,height=0.19\textheight]{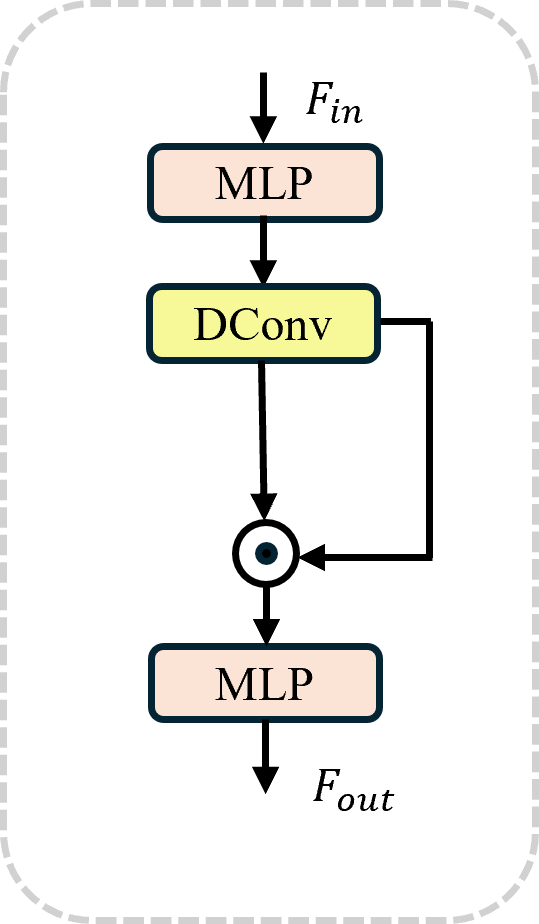}
\label{Fig 3c}}

\caption{Structural diagrams of each component. (a) Boundary Region Selective Scanning Module (BRSSM). (b) Global Selective Scanning Module (GSSM). (c) Structured Feed-Forward Network (SFFN)}
\label{Fig 3}
\end{figure*}

\textbf{BRSSM:} BRSSM integrates a boundary-region selective scanning mechanism that performs scanning on the feature map from the top, bottom, left, and right directions, focusing on the extraction of boundary information and regional features. Its core structure is based on the Visual State Space Block (VSSB) from Vmamba \cite{Liu2024VMambaVS}, with the detailed design shown in Figure \ref{Fig 3a}. The input first passes through a linear embedding layer and is then divided into two information streams. One stream passes through a 3$\times$3 depthwise separable convolution layer and a SiLU activation function before entering the core SSM. The output of the SSM is processed through layer normalization and then fused with the other stream, which is processed only through a SiLU activation function, using the Hadamard product. Finally, the fused information produces the final output.

\textbf{SFFN:} The proposed boundary-region selective scanning mechanism disrupts the original structural order of the image, limiting the model’s ability to capture the original structural information. To address this issue, the structural design of HomoFormer \cite{xiao2024homoformer} is adopted, shifting the structure modeling task to the FFN and implementing it as an Structured Feed-Forward Network (SFFN), as shown in Figure \ref{Fig 3c}. Specifically, a deep convolutional layer is inserted between two MLPs. By using local convolution operations, the model can assign weights based on the relative positions of features, enabling structure-aware interaction modeling and enhancing feature representation capability.

\subsubsection{\textbf{Global state-space block (GSSB)}}
GSSB shares the same structural design as BRSSB, with the only difference being that the BRSSM in BRSSB is replaced by the Global Selective Scanning Module (GSSM), as shown in the green part of Figure \ref{Fig 2}. Since the boundary-region selective scanning mechanism is prone to noise interference at low resolutions, and the features at this stage contain more high-level semantic information, GSSB is placed in the deeper layers of the U-Net. It uses global cross-scanning to directly extract brightness features, and restores the brightness of shadow regions using information from non-shadow areas.

\textbf{GSSM:} The GSSM incorporates a cross-scanning mechanism \cite{Liu2024VMambaVS} and follows the same structural design as BRSSM, as shown in Figure \ref{Fig 3b}. It scans the feature map from four directions: top, bottom, left, and right, allowing each element to gather global information from multiple directions. Consequently, the model is able to better capture brightness variations across directions and distinguish brightness differences between regions from multiple perspectives.

\subsection{Boundary-region selective scanning mechanism}

The boundary-region selective scanning mechanism is the core of the method proposed in this paper, specifically designed for image shadow removal tasks. When unfolding a 2D image into a 1D sequence, the arrangement of elements is crucial, as it directly affects the subsequent scanning process. Although Mamba can perform global modeling on long sequences, if semantically related pixels are far apart in the sequence, it weakens the model's performance. In other words, the closer the semantically related pixels are in the 1D sequence, the more beneficial it is for the representation ability of the state-space model. Therefore, reducing the distance between similar pixels in the long sequence helps enhance semantic continuity, thereby improving the performance of shadow removal.

\begin{figure*}[!t]
\centering

\subfloat[]{
		\includegraphics[width=0.22\textwidth,height=0.15\textheight]{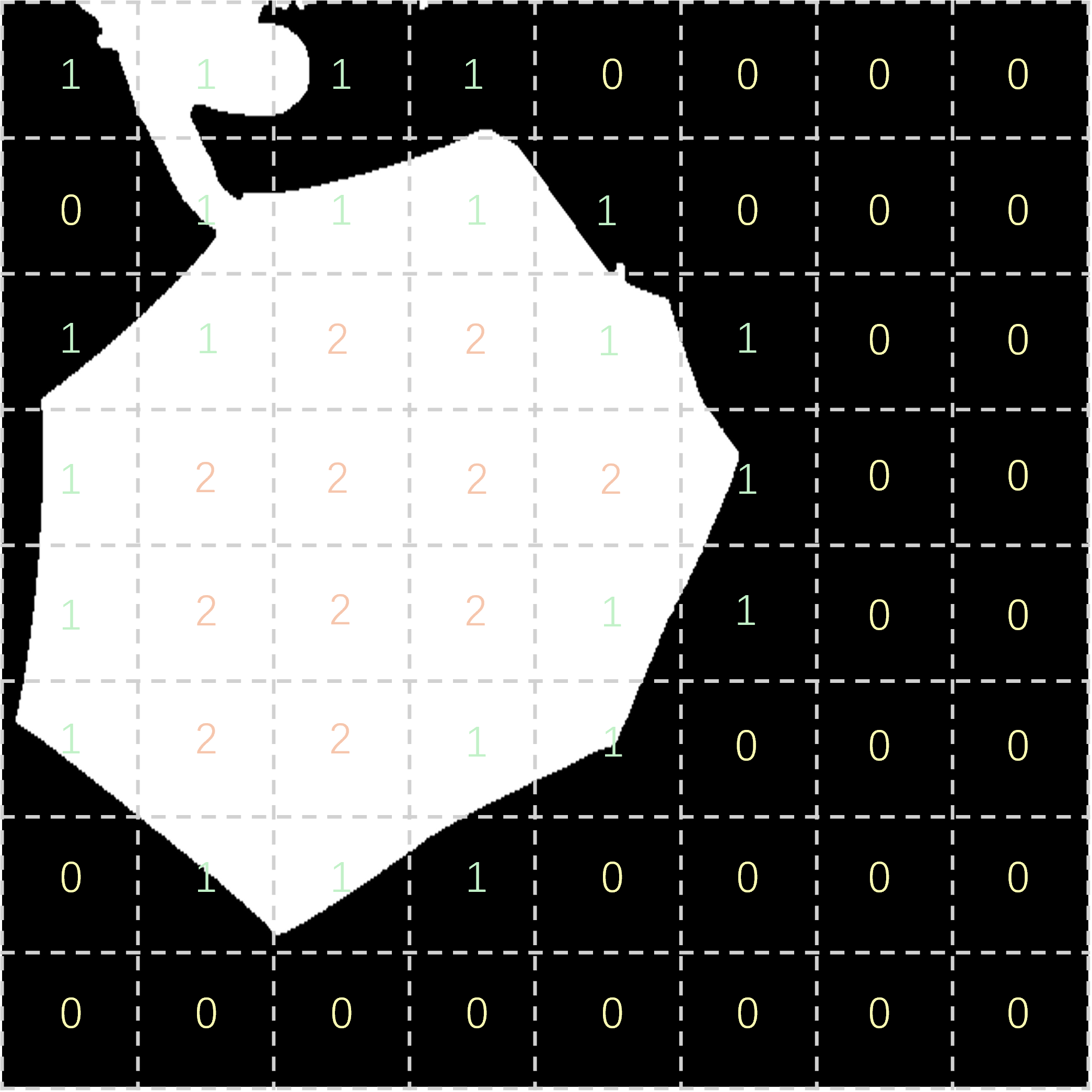}
\label{Fig 4a}}
\hspace{0.05cm}
\subfloat[]{
		\includegraphics[width=0.22\textwidth,height=0.15\textheight]{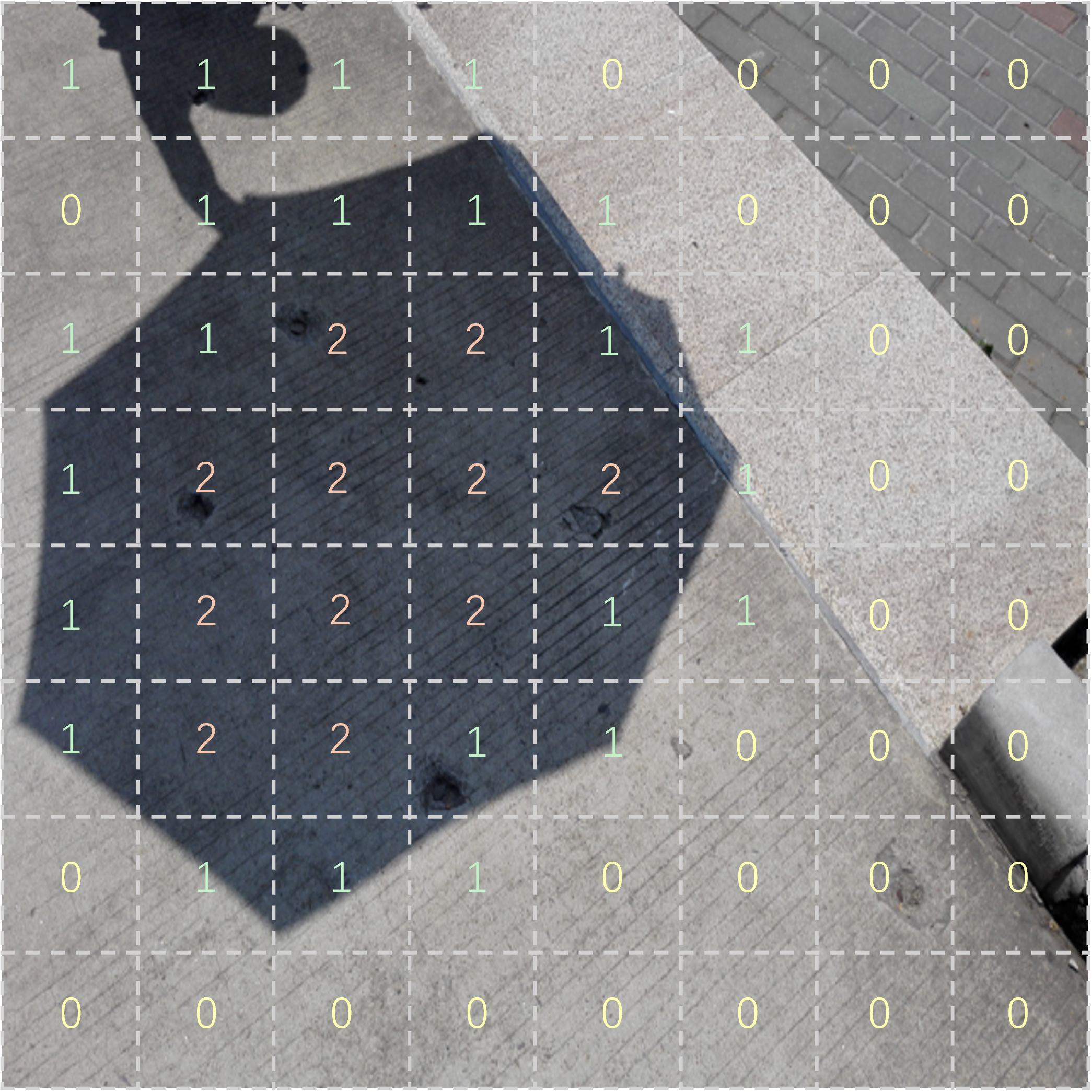}
\label{Fig 4b}}
\hspace{0.05cm}
\subfloat[]{
		\includegraphics[width=0.22\textwidth,height=0.15\textheight]{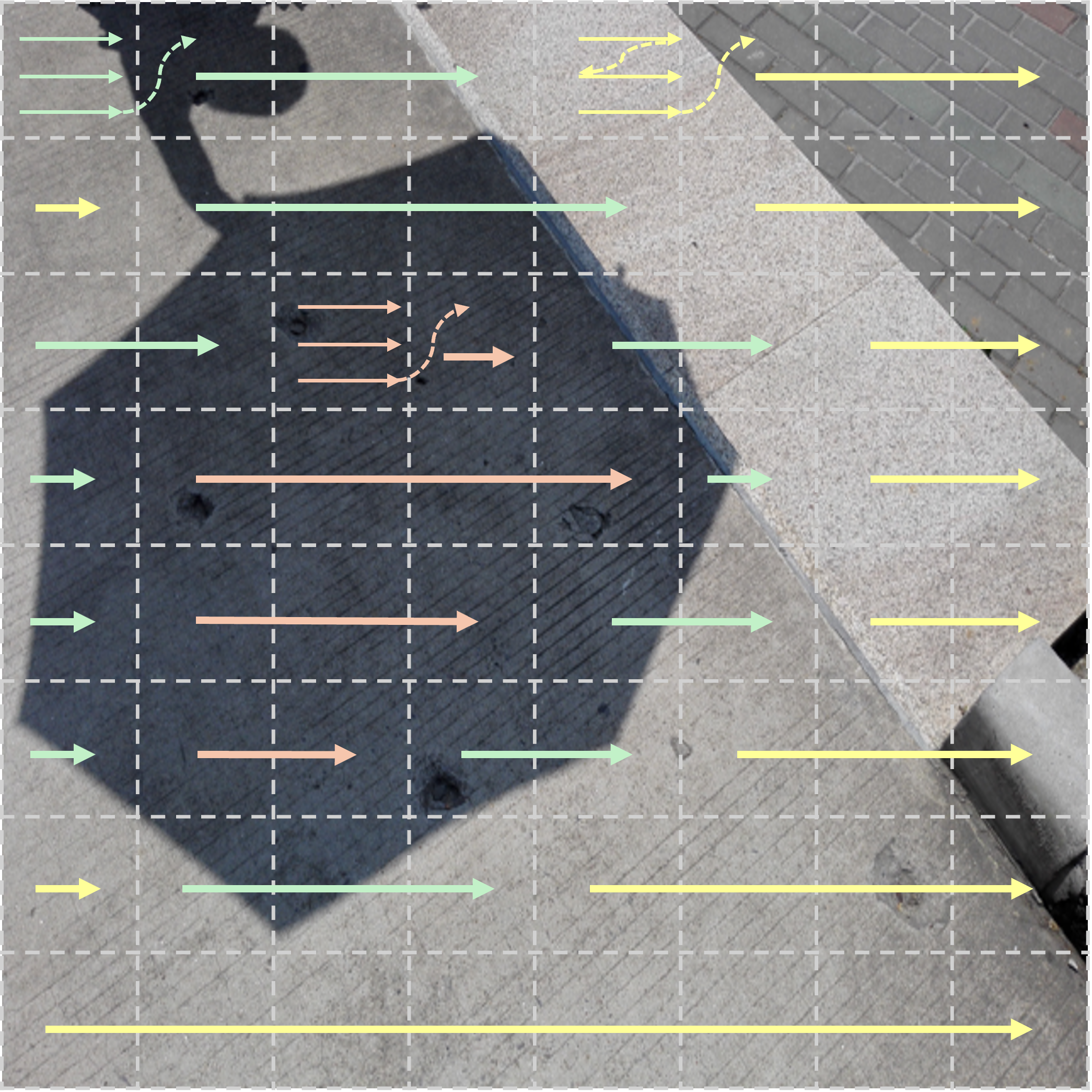}
\label{Fig 4c}}
\hspace{0.05cm}
\subfloat[]{
		\includegraphics[width=0.22\textwidth,height=0.15\textheight]{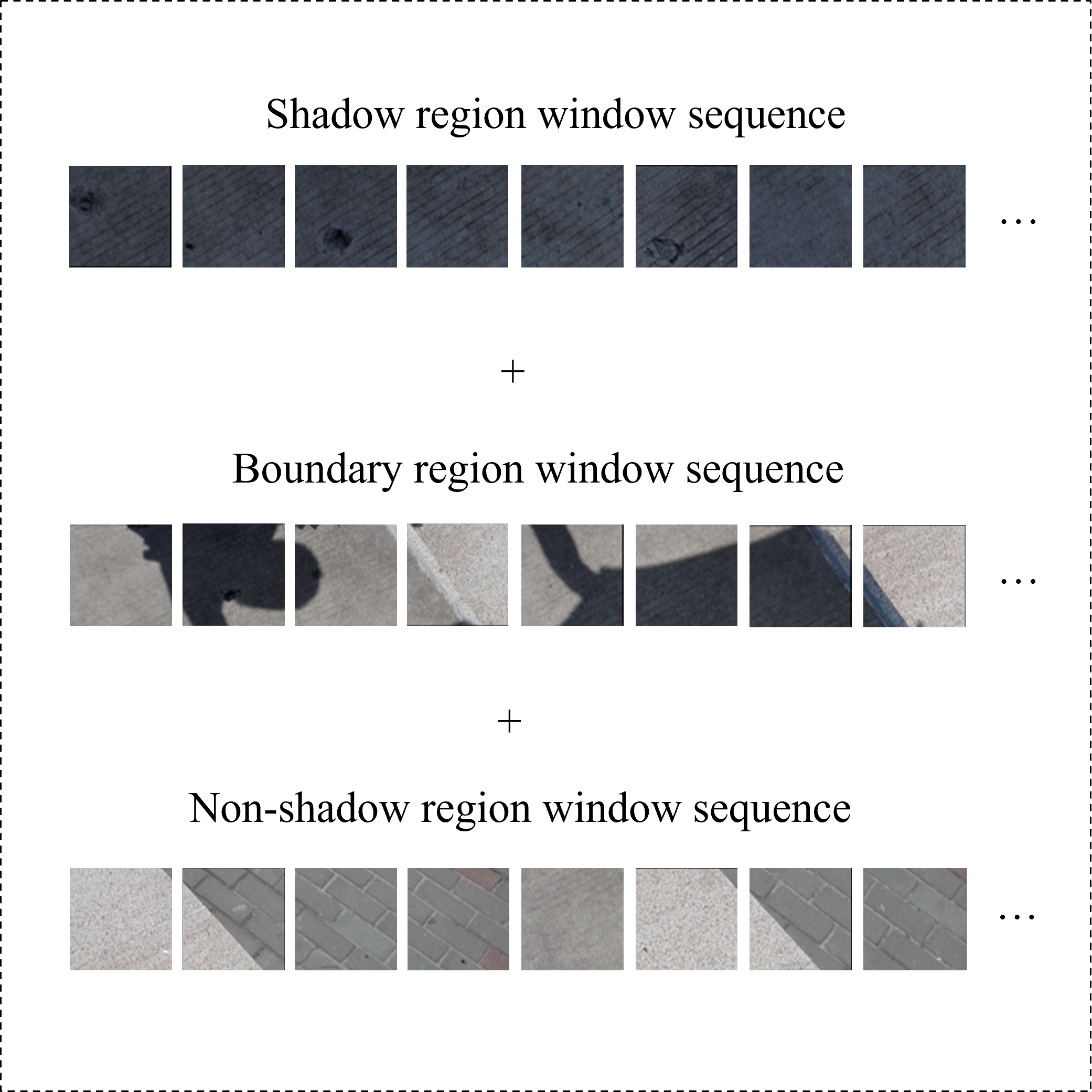}
\label{Fig 4d}}

\caption{Diagram of the principle of the boundary-region selective scanning mechanism. (a) Classify each window as belonging to the shadow region, non-shadow region, or boundary region. (b) The classified windows are mapped back to the original image. (c) Each type of window is scanned sequentially in horizontal and vertical directions. (d) Visualization of the scanning sequence.}
\label{Fig 4}
\end{figure*}

Based on the above theory, this paper proposes the boundary-region selective scanning mechanism. This mechanism uses windows as the basic unit for sorting and divides the image into shadow regions, non-shadow regions, and boundary regions using the shadow mask. Windows of the same type are grouped and sorted, and local scanning is then performed on all windows. This mechanism brings similar pixels closer together in the long sequence, enhances the semantic continuity of related pixels, and improves the model's understanding of local details. Figure \ref{Fig 4} illustrates the working principle of the boundary-region selective scanning mechanism.

Specifically, the proposed boundary-region selective scanning mechanism classifies windows based on the following rule: 
\begin{equation}
f(W) =
\begin{cases}
0, & \text{if } P(W) = \{0\}, \\
1, & \text{if } P(W) = \{0, 1\}, \\
2, & \text{if } P(W) = \{1\}.
\end{cases}
\end{equation}
$P(W)$ represents the set of pixel values in window $W$. When $P(W) = \{0\}$, all pixels in the window are 0, which belongs to category 0, indicating that the window is part of the non-shadow region. When $P(W) = \{0, 1\}$, the window contains both 0 and 1 (255), which belongs to category 1, indicating that the window is part of the boundary region. When $P(W) = \{1\}$, all elements in the window are 1 (255), which belongs to category 2, indicating that the window is part of the shadow region. 

This mechanism adopts four scanning directions: horizontal, vertical, reverse horizontal, and reverse vertical. A consistent scanning strategy is applied both within and between windows, allowing local information to be effectively aggregated from multiple directions. After processing with this mechanism, windows of the same category are arranged more closely in the 1D sequence, which effectively enhances the semantic continuity among semantically related pixels.

\subsection{Mask denoising method}

Models \cite{wan2024crformer, guo2023shadowformer} with mask-guided mechanisms typically rely on the accuracy of the mask, and ShadowMamba is no exception. When dealing with complex shadow images, the provided masks often contain noise, which interferes with the category classification process in the boundary-region selective scanning mechanism. Currently, this issue remains underexplored. Although conventional morphological operations are capable of removing noise, they often result in shadow boundary shifts, which may lead to more severe adverse effects. To address this, this paper proposes a simple and effective mask denoising method that removes most of the noise without shifting the boundaries. The method mainly targets noise in non-shadow regions, as these areas are relatively large and have a greater impact on overall modeling performance. After processing, the proposed boundary-region selective scanning mechanism is able to distinguish between shadow and non-shadow regions while preserving the internal semantic continuity within each region.

Specifically, the original mask is sequentially processed using opening and closing operations, resulting in a mask with noise removed but less accurate boundaries. The corresponding formulas are as follows:

\begin{align}
M_{\text{open}} &= \text{Dilation}(\text{Erosion}(M)) \\
M_{\text{close}} &= \text{Erosion}(\text{Dilation}(M_{\text{open}}))
\end{align}
Then, a dilation operation is applied to the processed mask to expand its coverage, resulting in a rough mask. The corresponding formula is shown below:
\begin{equation}
M_{\text{rough}} = \text{Dilation}(M_{\text{close}})
\end{equation}
Finally, the original mask is element-wise multiplied with the dilated rough mask to remove noise in the non-shadow regions. The specific formula is as follows:

\begin{equation}
M_{\text{denoised}} = M \cdot M_{\text{rough}}
\end{equation}
$M_{\text{denoised}}$ is the final denoised mask. 

It is worth noting that for models like ShadowFormer \cite{guo2023shadowformer}, which use mask-guided interaction in the deeper layers, this denoising method can significantly improve their performance. Its effectiveness is further verified in the following ablation studies.

\subsection{Loss function}
Only a single Charbonnier loss \cite{charbonnier1994two} is used in the proposed method to maintain pixel consistency, as shown in the following formula:

\begin{equation}
L(y, \hat{y}) = \sqrt{(y - \hat{y})^2 + \epsilon^2}
\end{equation}
where 
\( y \) is the ground truth shadow-free image,
\( \hat{y} \) is the predicted image output.
\( \epsilon \) is a small constant added to avoid numerical instability.

\section{Experiments}
\subsection{Implementation details}
The proposed ShadowMamba model is implemented using the PyTorch framework, and training is conducted on an RTX 4080 GPU. The network has a channel width of 32, and the batch size is set to 2. In its U-Net structure, the numbers of BRSSB and GSSB modules are [2, 2, 2, 2, 2, 2, 2]. The data augmentation techniques used include random cropping, rotation, flipping, and Mixup \cite{zhang2018mixup}. The AdamW optimizer is employed to update the learnable parameters, with the initial learning rate set to $2\times10^{-4}$, gradually decaying to $1\times10^{-6}$ using a cosine annealing strategy.

This study uses three commonly used datasets in the shadow removal field: ISTD \cite{wang2018stacked}, AISTD \cite{le2019shadow}, and SRD \cite{qu2017deshadownet}. The ISTD dataset contains 1,330 training samples and 540 testing samples, mainly consisting of hard shadows with clear boundaries. Each sample consists of a shadow image, a mask, and a shadow-free image, with the provided masks being nearly free of noise. Since there is a color difference between the shadow images and the ground truth images in the ISTD, AISTD corrects this difference using image processing algorithms. The SRD dataset includes both soft and hard shadows, with more complex backgrounds. It consists of 2,680 training samples and 408 testing samples. Each sample contains only a shadow image and a shadow-free image, with no masks provided. Following previous studies \cite{zhu2022bijective, guo2023shadowformer, mei2024latent}, this paper uses predicted masks generated by DHAN \cite{cun2020towardsss} during training and testing, which typically contain noise.

Similar to previous work \cite{le2021physics, einy2022physics, xiao2024homoformer, li2025shadowmaskformer}, the Root Mean Absolute Error (RMAE) in the Lab color space is used as an evaluation metric. Additionally, Peak Signal-to-Noise Ratio (PSNR) and Structural Similarity (SSIM) are employed to evaluate image performance in the RGB color space. For computational efficiency, FLOPs are measured on 256$\times$256 input images.

\subsection{Comparison with state-of-the-art methods} 

The proposed ShadowMamba is compared with the most popular state-of-the-art shadow removal models, including EMDN \cite{zhu2022efficient}, SG-ShadowNet \cite{wan2022style}, BA-ShadowNet \cite{niu2022boundary}, BM-Net \cite{zhu2022bijective}, Inpaint4shadow \cite{li2023leveraging}, ShadowFormer \cite{guo2023shadowformer}, ShadowDiffusion \cite{guo2023shadowdiffusion}, DMTN \cite{liu2023decoupled}, StructNet \cite{liu2023structure}, LFG-Diffusion \cite{mei2024latent}, DeS3 \cite{jin2024des3}, RRLSR \cite{liu2024recasting},  HomoFormer \cite{xiao2024homoformer}, OmniSR \cite{xu2025omnisr}, and ShadowMaskFormer \cite{li2025shadowmaskformer}. To ensure a fair comparison, the results of these compared methods are obtained from the original papers.

\subsubsection{\textbf{Quantitative measure}}Table \ref{Tab 1}, \ref{Tab 2}\&\ref{Tab 3} show the quantitative results on the testing sets over AISTD, SRD, and ISTD respectively. The results show that the proposed method achieves superior performance on both hard shadow and soft shadow datasets.

\begin{table*}[]
\centering

\caption{Quantitative comparisons with the SOTA methods on the AISTD dataset \cite{le2019shadow}.}
\label{Tab 1}

\begin{tabular}{c|c|ccc|ccc|ccc}
\hline
\multirow{2}{*}{Size}        & \multirow{2}{*}{Method} & \multicolumn{3}{c|}{All Image (ALL)}                                             & \multicolumn{3}{c|}{Shadow Region (S)}                                           & \multicolumn{3}{c}{Non-Shadow Region (NS)}                                       \\
\multicolumn{1}{l|}{}                  &                             & \multicolumn{1}{c}{PSNR$\uparrow$}           & \multicolumn{1}{c}{SSIM$\uparrow$}           & RMAE$\downarrow$          & \multicolumn{1}{c}{PSNR$\uparrow$}           & \multicolumn{1}{c}{SSIM$\uparrow$}           & RMAE$\downarrow$         & \multicolumn{1}{c}{PSNR$\uparrow$}           & \multicolumn{1}{c}{SSIM$\uparrow$}           & RMAE$\downarrow$          \\ \hline
\multirow{9}{*}{\rotatebox{90}{256$\times$256}}               & BA-ShadowNet \cite{niu2022boundary}                     & \multicolumn{1}{c|}{34.60}          & \multicolumn{1}{c|}{0.971}          & 3.00          & \multicolumn{1}{c|}{38.29}          & \multicolumn{1}{c|}{0.991}          & 5.90          & \multicolumn{1}{c|}{38.48}          & \multicolumn{1}{c|}{0.984}          & 3.25          \\ 
                                       & Inpaint4Shadow \cite{li2023leveraging}              & \multicolumn{1}{c|}{34.16}          & \multicolumn{1}{c|}{0.967}          & 3.35          & \multicolumn{1}{c|}{38.10}          & \multicolumn{1}{c|}{0.990}          & 6.09          & \multicolumn{1}{c|}{37.66}          & \multicolumn{1}{c|}{0.981}          & 2.82          \\
                                       & ShadowDiffusion \cite{guo2023shadowdiffusion}             & \multicolumn{1}{c|}{\underline{35.67}}          & \multicolumn{1}{c|}{\textbf{0.975}}          & 2.72          & \multicolumn{1}{c|}{\underline{39.69}}          & \multicolumn{1}{c|}{0.992}          & 4.97          & \multicolumn{1}{c|}{38.89}          & \multicolumn{1}{c|}{\textbf{0.987}} & 2.28          \\

 & StructNet \cite{liu2023structure}             & \multicolumn{1}{c|}{34.26}          & \multicolumn{1}{c|}{0.969}          & -          & \multicolumn{1}{c|}{37.92}          & \multicolumn{1}{c|}{0.991}          & -          & \multicolumn{1}{c|}{37.72}          & \multicolumn{1}{c|}{0.983} & - \\

                                       & DeS3 \cite{jin2024des3}                       & \multicolumn{1}{c|}{31.38}          & \multicolumn{1}{c|}{0.957}          & 3.85          & \multicolumn{1}{c|}{36.49}          & \multicolumn{1}{c|}{0.989}          & 6.59          & \multicolumn{1}{c|}{34.72}          & \multicolumn{1}{c|}{0.972}          & 3.32          \\
                                       & RRLSR \cite{liu2024recasting}                       & \multicolumn{1}{c|}{34.96}          & \multicolumn{1}{c|}{0.968}          & 2.87          & \multicolumn{1}{c|}{38.04}          & \multicolumn{1}{c|}{0.990}          & 5.69          & \multicolumn{1}{c|}{\textbf{39.15}}          & \multicolumn{1}{c|}{0.984}          & 2.31          \\
                                       & HomoFormer \cite{xiao2024homoformer}                 & \multicolumn{1}{c|}{35.35}          & \multicolumn{1}{c|}{\textbf{0.975}}          & \underline{2.64}          & \multicolumn{1}{c|}{39.49}          & \multicolumn{1}{c|}{\textbf{0.993}}          & \textbf{4.73} & \multicolumn{1}{c|}{38.75}          & \multicolumn{1}{c|}{0.984}          & \underline{2.23}          \\ \cline{2-11} 
                                       & \textbf{ShadowMamba} & \multicolumn{1}{c|}{\textbf{35.74}} & \multicolumn{1}{c|}{\textbf{0.975}} & \textbf{2.59} & \multicolumn{1}{c|}{\textbf{39.96}} & \multicolumn{1}{c|}{\textbf{0.993}} & \underline{4.77}          & \multicolumn{1}{c|}{\underline{39.05}} & \multicolumn{1}{c|}{\underline{0.986}}          & \textbf{2.17} \\ \hline
\multirow{6}{*}{\rotatebox{90}{640$\times480$}}              & SG-ShadowNet \cite{wan2022style}               & \multicolumn{1}{c|}{30.50}          & \multicolumn{1}{c|}{0.930}          & 4.28          & \multicolumn{1}{c|}{35.96}          & \multicolumn{1}{c|}{0.984}          & 6.73          & \multicolumn{1}{c|}{32.76}          & \multicolumn{1}{c|}{0.950}          & 3.82          \\ 
                                       & BM-Net \cite{zhu2022bijective}                     & \multicolumn{1}{c|}{31.85}          & \multicolumn{1}{c|}{\underline{0.941}}          & 3.73          & \multicolumn{1}{c|}{36.80}          & \multicolumn{1}{c|}{\textbf{0.987}}          & 6.32          & \multicolumn{1}{c|}{34.47}          & \multicolumn{1}{c|}{\textbf{0.958}}          & 3.25          \\ 
                                       & ShadowFormer \cite{guo2023shadowformer}                & \multicolumn{1}{c|}{\underline{32.78}}          & \multicolumn{1}{c|}{0.939}          & \underline{3.61}          & \multicolumn{1}{c|}{\underline{38.07}}          & \multicolumn{1}{c|}{0.986}          & 6.02          & \multicolumn{1}{c|}{\textbf{35.14}}          & \multicolumn{1}{c|}{0.955}          & \underline{3.15}          \\ 
                                       & LFG-Diffusion \cite{mei2024latent}              & \multicolumn{1}{c|}{32.11}          & \multicolumn{1}{c|}{0.936}          & 3.89          & \multicolumn{1}{c|}{37.74}          & \multicolumn{1}{c|}{\textbf{0.987}}          & \underline{5.91}          & \multicolumn{1}{c|}{34.30}          & \multicolumn{1}{c|}{0.951}          & 3.51          \\ 
                                       & OmniSR \cite{xu2025omnisr}                     & \multicolumn{1}{c|}{31.35}          & \multicolumn{1}{c|}{0.939}          & 3.87          & \multicolumn{1}{c|}{36.16}          & \multicolumn{1}{c|}{0.986}          & 7.18          & \multicolumn{1}{c|}{34.41}          & \multicolumn{1}{c|}{0.956}          & 3.25          \\ \cline{2-11} 
                                       & \textbf{ShadowMamba} & \multicolumn{1}{c|}{\textbf{32.85}} & \multicolumn{1}{c|}{\textbf{0.943}} & \textbf{3.42} & \multicolumn{1}{c|}{\textbf{38.43}} & \multicolumn{1}{c|}{\textbf{0.987}} & \textbf{5.51} & \multicolumn{1}{c|}{\underline{35.09}} & \multicolumn{1}{c|}{\textbf{0.958}} & \textbf{3.03} \\ \hline
\end{tabular}
\end{table*}

\begin{table*}[]
\centering
\caption{Quantitative comparisons with the SOTA methods on the SRD dataset \cite{qu2017deshadownet}.}
\label{Tab 2}

\begin{tabular}{c|c|ccc|ccc|ccc}
\hline
\multirow{2}{*}{Size}        & \multirow{2}{*}{Method} & \multicolumn{3}{c|}{All Image (ALL)}                                             & \multicolumn{3}{c|}{Shadow Region (S)}                                           & \multicolumn{3}{c}{Non-Shadow Region (NS)}                                       \\
\multicolumn{1}{l|}{}                  &                             & \multicolumn{1}{c}{PSNR$\uparrow$}           & \multicolumn{1}{c}{SSIM$\uparrow$}           & RMAE$\downarrow$          & \multicolumn{1}{c}{PSNR$\uparrow$}           & \multicolumn{1}{c}{SSIM$\uparrow$}           & RMAE$\downarrow$         & \multicolumn{1}{c}{PSNR$\uparrow$}           & \multicolumn{1}{c}{SSIM$\uparrow$}           & RMAE$\downarrow$          \\ \hline
\multirow{10}{*}{\rotatebox{90}{256$\times$256}}                 & EMDN \cite{zhu2022efficient}                            & \multicolumn{1}{c|}{31.72}          & \multicolumn{1}{c|}{0.952}          & 4.79          & \multicolumn{1}{c|}{34.94}          & \multicolumn{1}{c|}{0.980}          & 7.44          & \multicolumn{1}{c|}{35.85}          & \multicolumn{1}{c|}{0.982}          & 3.74          \\
                          & BM-Net \cite{zhu2022bijective}                          & \multicolumn{1}{c|}{31.69}          & \multicolumn{1}{c|}{0.956}          & 4.46          & \multicolumn{1}{c|}{35.05}          & \multicolumn{1}{c|}{0.981}          & 6.61          & \multicolumn{1}{c|}{36.02}          & \multicolumn{1}{c|}{0.982}          & 3.61          \\
                          & Inpaint4Shadow \cite{li2023leveraging}                 & \multicolumn{1}{c|}{33.27}          & \multicolumn{1}{c|}{0.967}          & 3.81          & \multicolumn{1}{c|}{36.73}          & \multicolumn{1}{c|}{0.985}          & 5.70          & \multicolumn{1}{c|}{36.70}          & \multicolumn{1}{c|}{0.985}          & 3.27          \\
                          & ShadowFormer \cite{guo2023shadowformer}                    & \multicolumn{1}{c|}{32.46}          & \multicolumn{1}{c|}{0.957}          & 4.28          & \multicolumn{1}{c|}{35.55}          & \multicolumn{1}{c|}{0.982}          & 6.14          & \multicolumn{1}{c|}{36.82}          & \multicolumn{1}{c|}{0.983}          & 3.54          \\
                          & ShadowDiffusion \cite{guo2023shadowdiffusion}                 & \multicolumn{1}{c|}{34.73}          & \multicolumn{1}{c|}{0.970}          & 3.63          & \multicolumn{1}{c|}{38.72}          & \multicolumn{1}{c|}{\underline{0.987}}          & 4.98          & \multicolumn{1}{c|}{37.78}          & \multicolumn{1}{c|}{0.985}          & 3.44          \\
                          & DeS3 \cite{jin2024des3}                           & \multicolumn{1}{c|}{34.11}          & \multicolumn{1}{c|}{0.968}          & 3.56          & \multicolumn{1}{c|}{37.91}          & \multicolumn{1}{c|}{0.986}          & 5.27          & \multicolumn{1}{c|}{37.45}          & \multicolumn{1}{c|}{0.984}          & 3.03          \\
                          & OmniSR \cite{xu2025omnisr}                         & \multicolumn{1}{c|}{32.87}          & \multicolumn{1}{c|}{0.969}          & -             & \multicolumn{1}{c|}{-}              & \multicolumn{1}{c|}{-}              & -             & \multicolumn{1}{c|}{-}              & \multicolumn{1}{c|}{-}              & -             \\
                          & HomoFormer \cite{xiao2024homoformer}                      & \multicolumn{1}{c|}{\textbf{35.37}} & \multicolumn{1}{c|}{\underline{0.972}}          & \underline{3.33}          & \multicolumn{1}{c|}{\underline{38.81}}          & \multicolumn{1}{c|}{\underline{0.987}}          & \textbf{4.25} & \multicolumn{1}{c|}{\textbf{39.45}} & \multicolumn{1}{c|}{\underline{0.988}}          & \underline{2.85}          \\ \cline{2-11} 
                          & \textbf{ShadowMamba}     & \multicolumn{1}{c|}{\underline{35.32}}          & \multicolumn{1}{c|}{\textbf{0.980}} & \textbf{3.25} & \multicolumn{1}{c|}{\textbf{39.07}} & \multicolumn{1}{c|}{\textbf{0.990}} & \underline{4.85}          & \multicolumn{1}{c|}{\underline{39.25}}          & \multicolumn{1}{c|}{\textbf{0.993}} & \textbf{2.64} \\ \hline
\end{tabular}
\end{table*}

\begin{table*}[]
\centering

\caption{Quantitative comparisons with the SOTA methods on the ISTD dataset \cite{wang2018stacked}.}
\label{Tab 3}

\begin{tabular}{c|c|ccc|ccc|ccc}
\hline
\multirow{2}{*}{Size}        & \multirow{2}{*}{Method} & \multicolumn{3}{c|}{All Image (ALL)}                                             & \multicolumn{3}{c|}{Shadow Region (S)}                                           & \multicolumn{3}{c}{Non-Shadow Region (NS)}                                       \\
\multicolumn{1}{l|}{}                  &                             & \multicolumn{1}{c}{PSNR$\uparrow$}           & \multicolumn{1}{c}{SSIM$\uparrow$}           & RMAE$\downarrow$          & \multicolumn{1}{c}{PSNR$\uparrow$}           & \multicolumn{1}{c}{SSIM$\uparrow$}           & RMAE$\downarrow$         & \multicolumn{1}{c}{PSNR$\uparrow$}           & \multicolumn{1}{c}{SSIM$\uparrow$}           & RMAE$\downarrow$          \\ \hline
\multirow{8}{*}{\rotatebox{90}{256$\times$256}} 
                         & BM-Net \cite{zhu2022bijective}                          & \multicolumn{1}{c|}{30.28}          & \multicolumn{1}{c|}{0.959}          & 5.02          & \multicolumn{1}{c|}{35.61}          & \multicolumn{1}{c|}{0.988}          & 7.34          & \multicolumn{1}{c|}{32.80}          & \multicolumn{1}{c|}{0.976}          & 4.57          \\
                         & EMDN \cite{zhu2022efficient}                           & \multicolumn{1}{c|}{29.98}          & \multicolumn{1}{c|}{0.944}          & 5.22          & \multicolumn{1}{c|}{36.27}          & \multicolumn{1}{c|}{0.986}          & 7.78          & \multicolumn{1}{c|}{31.85}          & \multicolumn{1}{c|}{0.965}          & 4.72          \\
 & StructNet \cite{liu2023structure}             & \multicolumn{1}{c|}{30.32}          & \multicolumn{1}{c|}{0.963}          & -          & \multicolumn{1}{c|}{36.40}          & \multicolumn{1}{c|}{0.989}          & -          & \multicolumn{1}{c|}{32.27}          & \multicolumn{1}{c|}{0.978} & - \\

 & DMTN \cite{liu2023decoupled}             & \multicolumn{1}{c|}{30.42}          & \multicolumn{1}{c|}{0.965}          & -          & \multicolumn{1}{c|}{35.83}          & \multicolumn{1}{c|}{0.990}          & -          & \multicolumn{1}{c|}{33.01}          & \multicolumn{1}{c|}{0.979} & - \\

                         & ShadowFormer \cite{guo2023shadowformer}                   & \multicolumn{1}{c|}{\underline{32.21}}          & \multicolumn{1}{c|}{\underline{0.968}}          & \underline{4.09}          & \multicolumn{1}{c|}{\underline{38.19}}          & \multicolumn{1}{c|}{\textbf{0.991}}          & \underline{5.96}          & \multicolumn{1}{c|}{\underline{34.32}}          & \multicolumn{1}{c|}{\textbf{0.981}}          & \underline{3.72}          \\
                         & ShadowMaskFormer \cite{li2025shadowmaskformer}                   & \multicolumn{1}{c|}{-}          & \multicolumn{1}{c|}{-}          & 4.23          & \multicolumn{1}{c|}{-}          & \multicolumn{1}{c|}{-}          & 6.08          & \multicolumn{1}{c|}{-}          & \multicolumn{1}{c|}{-}          & 3.86        \\ \cline{2-11} 
                         & \textbf{ShadowMamba}     & \multicolumn{1}{c|}{\textbf{32.78}} & \multicolumn{1}{c|}{\textbf{0.970}} & \textbf{4.06} & \multicolumn{1}{c|}{\textbf{38.90}}          & \multicolumn{1}{c|}{\textbf{0.991}}          &\textbf{5.83}          & \multicolumn{1}{c|}{\textbf{34.97}} & \multicolumn{1}{c|}{\textbf{0.981}} & \textbf{3.71} \\ \hline
\end{tabular}
\end{table*}

On the hard shadow datasets AISTD \cite{le2019shadow} and ISTD \cite{wang2018stacked}, the boundary-region selective scanning mechanism fully exploits its advantages due to the provided masks having clear boundaries with minimal noise. Methods such as ShadowFormer \cite{guo2023shadowformer}, OmniSR \cite{xu2025omnisr}, and HomoFormer \cite{xiao2024homoformer} are limited by the size of the local window, making it difficult to achieve true global modeling. In contrast, ShadowMamba is able to model global tokens with linear complexity, resulting in superior performance. Moreover, ShadowMamba outperforms diffusion-based methods such as ShadowDiffusion \cite{guo2023shadowdiffusion}, LFG-Diffusion \cite{mei2024latent}, and RRLSR \cite{liu2024recasting}. Although diffusion models have certain advantages in generation quality, their inference speed is relatively slow. In contrast, ShadowMamba achieves higher inference efficiency with a lower parameter count.

On the soft shadow dataset SRD \cite{qu2017deshadownet}, ShadowMamba achieves performance comparable to the SOTA model HomoFormer \cite{xiao2024homoformer} and significantly outperforms DeS3 \cite{jin2024des3}, which is specifically designed for soft shadow removal. Its boundary-region selective scanning mechanism uses windows as basic units and can effectively extract local boundary information, even when handling soft shadows with blurred edges and smooth transitions. In addition, the mask denoising method significantly reduces noise interference, ensuring effective modeling of this mechanism.

Table~\ref{Tab 4} presents a comparison of the models in terms of computational complexity and the number of parameters. ShadowMamba significantly reduces the number of parameters, mainly due to its hierarchical combination strategy, where shallow and deep layers take on different roles instead of combining both local and global operations within a single block. This architecture helps maintain the model's representation ability while effectively controlling the overall parameter count. In addition, ShadowMamba also has slightly lower computational complexity than models \cite{guo2023shadowformer, xiao2024homoformer} based on window attention mechanisms, further highlighting its efficiency while maintaining strong performance.

\begin{table}[]
\centering
\caption{Quantitative comparison of the methods' computational efficiency in terms of the number of trainable parameters (in million, M) and FLOPs (in billion, G).}
\label{Tab 4}

\begin{tabular}{c|c|c|c}
\hline
 Size                   & Method            & Params (M)$\downarrow$ & FLOPs (G)$\downarrow$ \\ \hline
\multirow{6}{*}{\rotatebox{90}{256$\times$256}} & ShadowDiffusion \cite{guo2023shadowdiffusion}           & 60.74               & 937.15             \\
                     & Inpaint4Shadow \cite{li2023leveraging}             & 14.98               & 81.18              \\
				& StructNet \cite{liu2023structure}                 & 67.06              & 45.95              \\
                     & ShadowFormer \cite{guo2023shadowformer}              & \underline{11.35}               & 64.60              \\
                     & HomoFormer \cite{xiao2024homoformer}                & 17.81               & \underline{35.63}              \\ \cline{2-4} 
                     & \textbf{ShadowMamba} & \textbf{6.45}                & \textbf{34.71}              \\ \hline
\end{tabular}
\end{table}

\subsubsection{\textbf{Qualitative measure}}
\begin{figure*}[htbp]
    \centering

    \begin{subfigure}[b]{0.115\linewidth}
        \centering
        \subcaption*{Input} 
        \includegraphics[width=2cm,height=1.5cm]{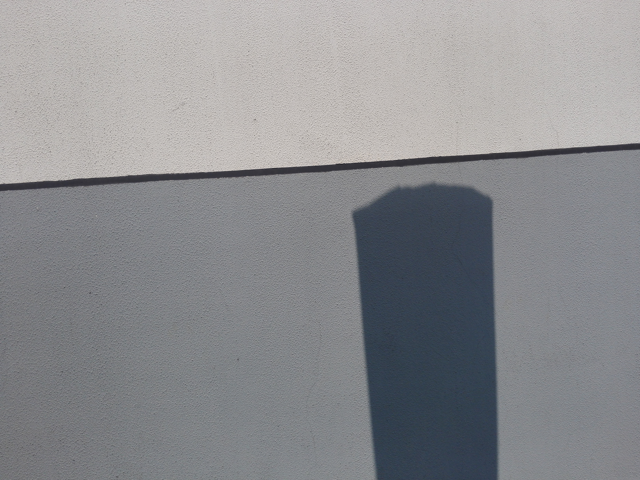}\\[0.05cm]
		\includegraphics[width=2cm,height=1.5cm]{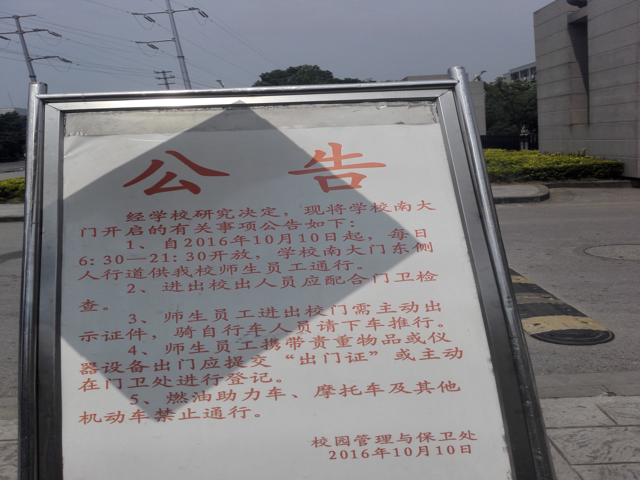}
    \end{subfigure}
    \begin{subfigure}[b]{0.115\linewidth}
        \centering
        \subcaption*{BM-Net \cite{zhu2022bijective}}
        \includegraphics[width=2cm,height=1.5cm]{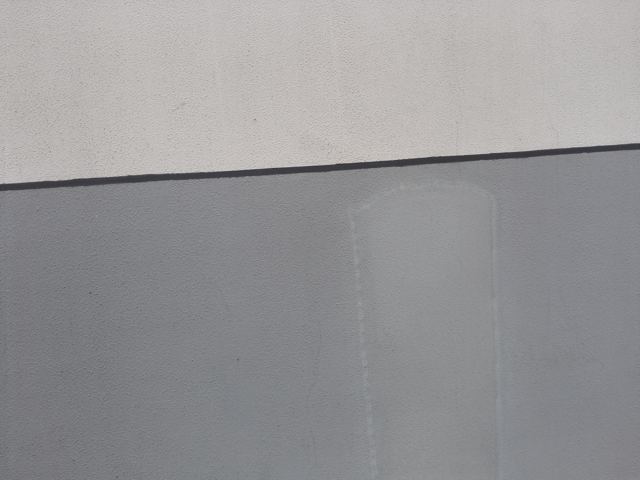}\\[0.05cm]
        \includegraphics[width=2cm,height=1.5cm]{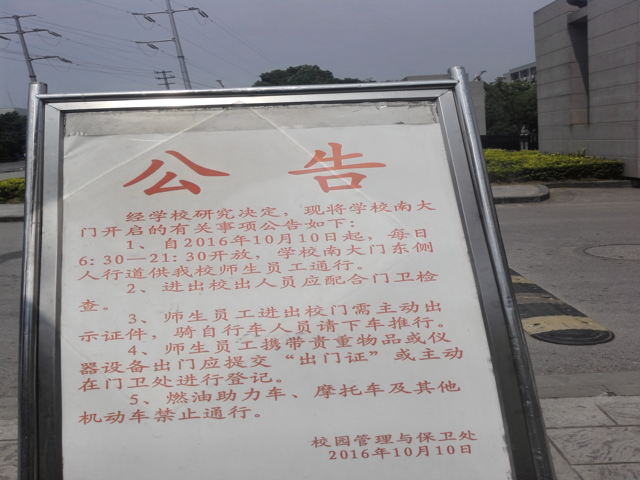}
    \end{subfigure}
    \begin{subfigure}[b]{0.115\linewidth}
        \centering
        \subcaption*{Inpaint4s \cite{li2023leveraging}}
        \includegraphics[width=2cm,height=1.5cm]{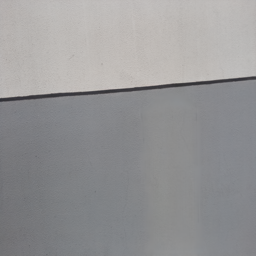}\\[0.05cm]
        \includegraphics[width=2cm,height=1.5cm]{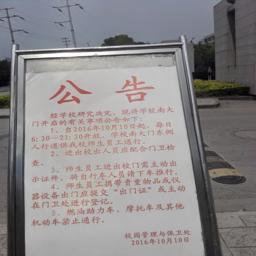}
    \end{subfigure}
    \begin{subfigure}[b]{0.115\linewidth}
        \centering
        \subcaption*{SF \cite{guo2023shadowformer}}
        \includegraphics[width=2cm,height=1.5cm]{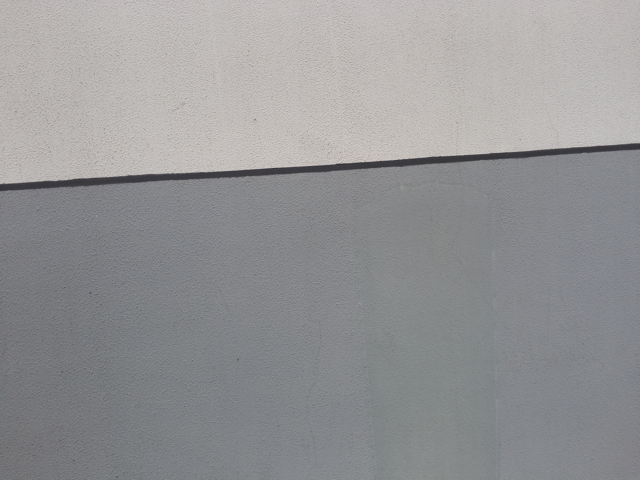}\\[0.05cm]
        \includegraphics[width=2cm,height=1.5cm]{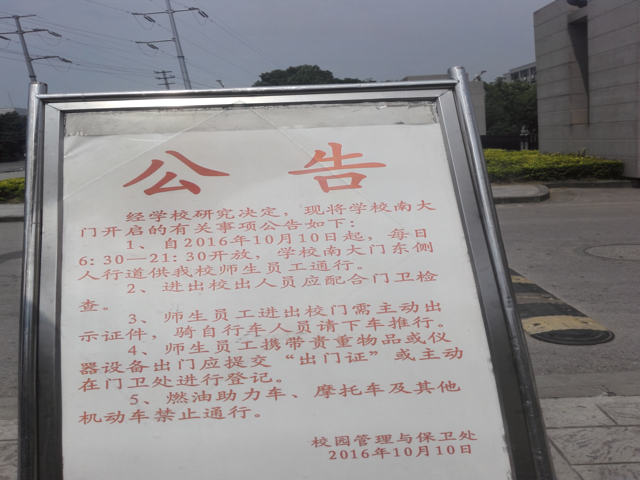}
    \end{subfigure}
    \begin{subfigure}[b]{0.115\linewidth}
        \centering
        \subcaption*{LFG-Diff \cite{mei2024latent}}
        \includegraphics[width=2cm,height=1.5cm]{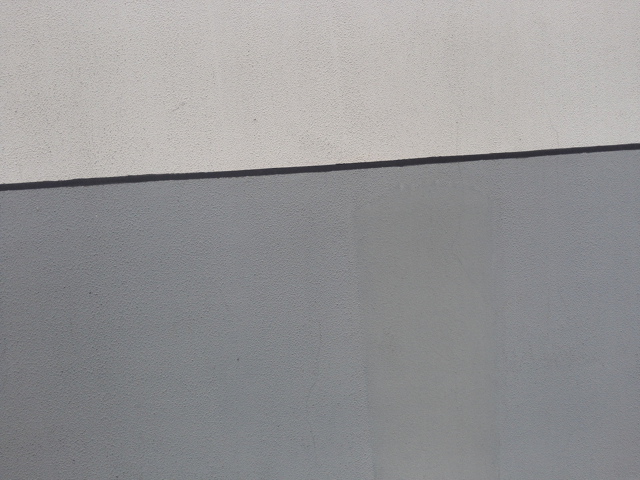}\\[0.05cm]
        \includegraphics[width=2cm,height=1.5cm]{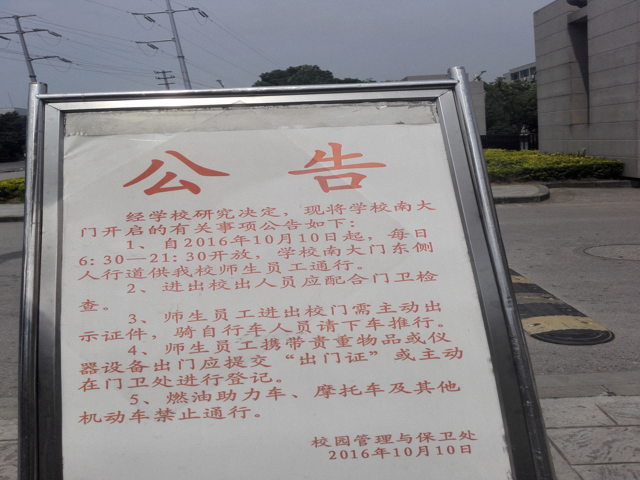}
    \end{subfigure}
    \begin{subfigure}[b]{0.115\linewidth}
        \centering
        \subcaption*{SDiff \cite{guo2023shadowdiffusion}}
        \includegraphics[width=2cm,height=1.5cm]{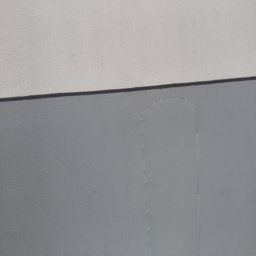}\\[0.05cm]
        \includegraphics[width=2cm,height=1.5cm]{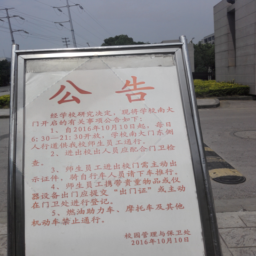}
    \end{subfigure}
    \begin{subfigure}[b]{0.115\linewidth}
        \centering
        \subcaption*{\textbf{Ours}}
        \includegraphics[width=2cm,height=1.5cm]{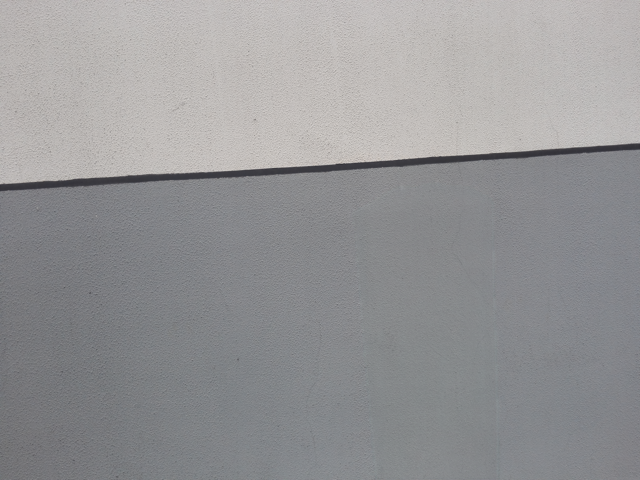}\\[0.05cm]
        \includegraphics[width=2cm,height=1.5cm]{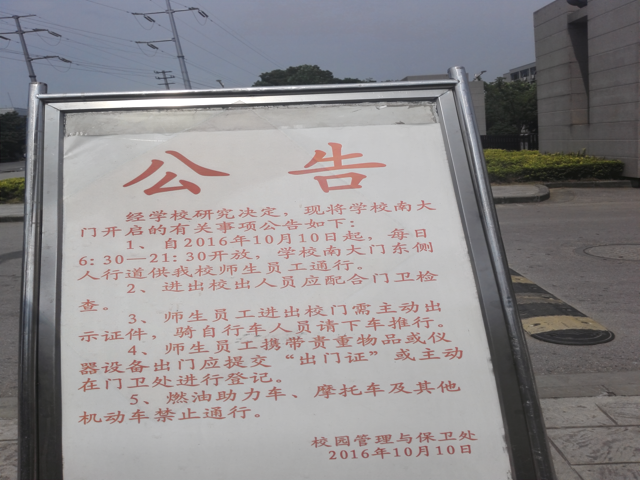}
    \end{subfigure}
    \begin{subfigure}[b]{0.115\linewidth}
        \centering
        \subcaption*{Ground Truth}
        \includegraphics[width=2cm,height=1.5cm]{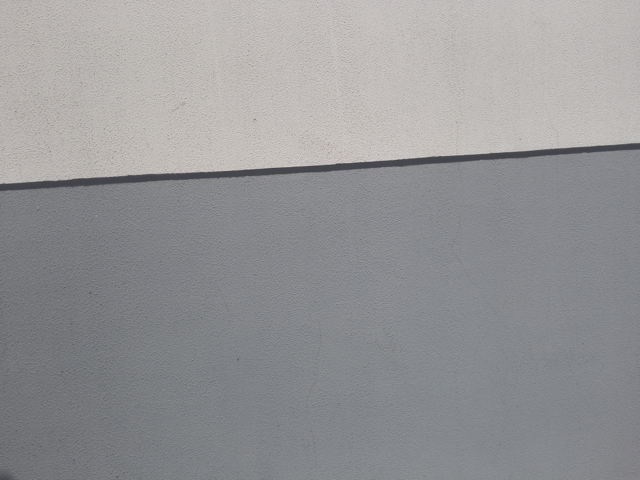}\\[0.05cm]
        \includegraphics[width=2cm,height=1.5cm]{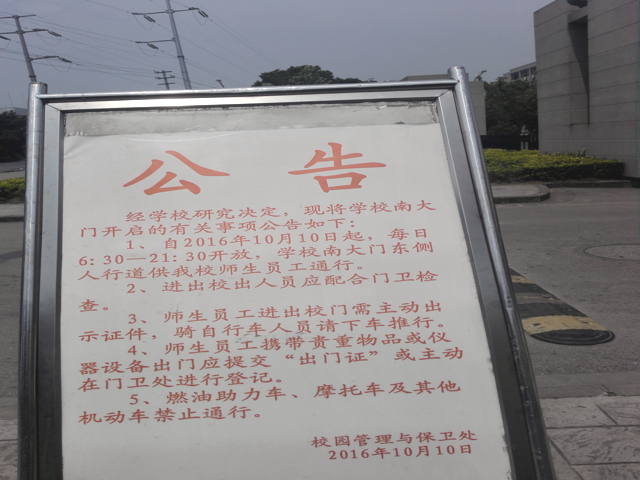}
    \end{subfigure}

    \caption{Qualitative results comparing ShadowMamba with other methods on the AISTD dataset (model names abbreviated).} 
    \label{Fig 6}
\end{figure*}

\begin{figure*}[htbp]
    \centering

    \begin{subfigure}[b]{0.115\linewidth}
        \centering
        \subcaption*{Input} 
        \includegraphics[width=2cm,height=1.5cm]{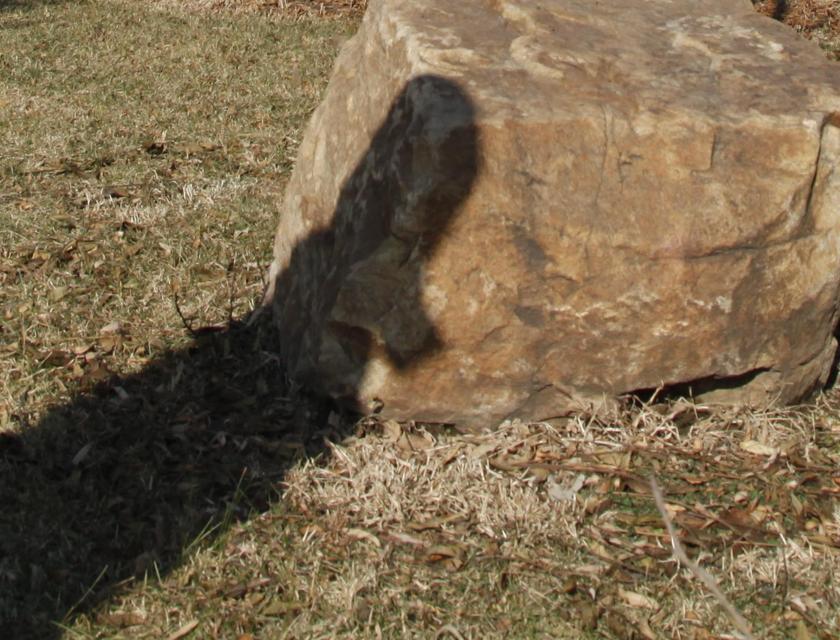}\\[0.05cm]
        \includegraphics[width=2cm,height=1.5cm]{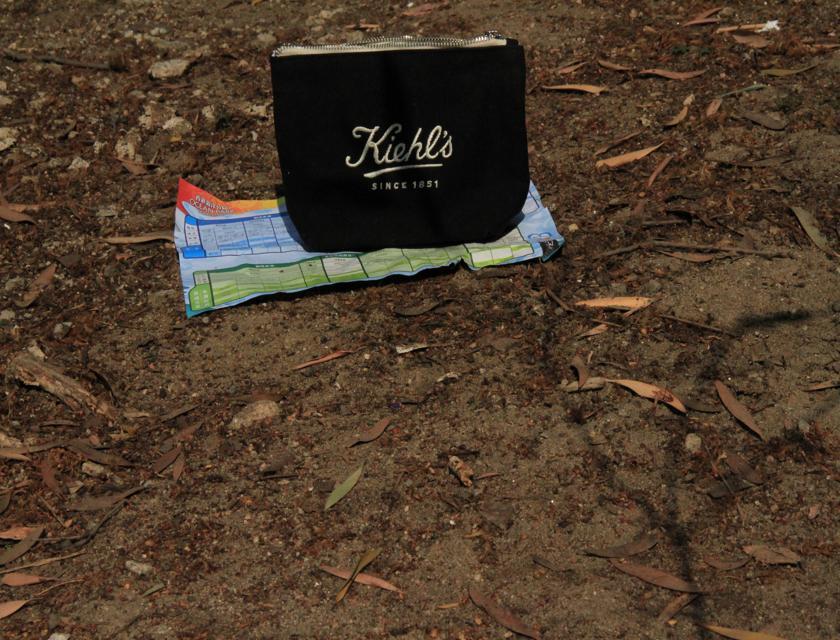}
    \end{subfigure}
    \begin{subfigure}[b]{0.115\linewidth}
        \centering
        \subcaption*{SG \cite{wan2022style}}
        \includegraphics[width=2cm,height=1.5cm]{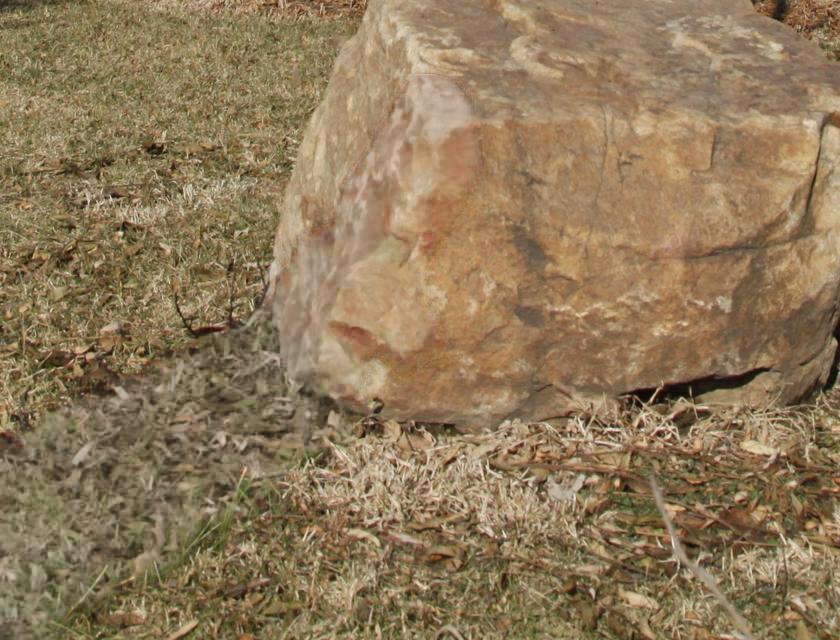}\\[0.05cm]
        \includegraphics[width=2cm,height=1.5cm]{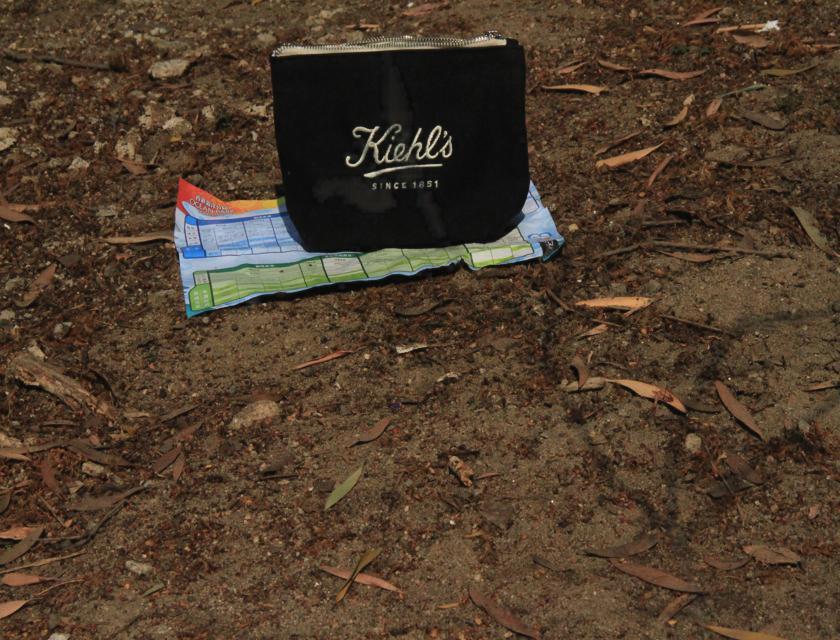}
    \end{subfigure}
    \begin{subfigure}[b]{0.115\linewidth}
        \centering
        \subcaption*{BM-Net \cite{zhu2022bijective}}
        \includegraphics[width=2cm,height=1.5cm]{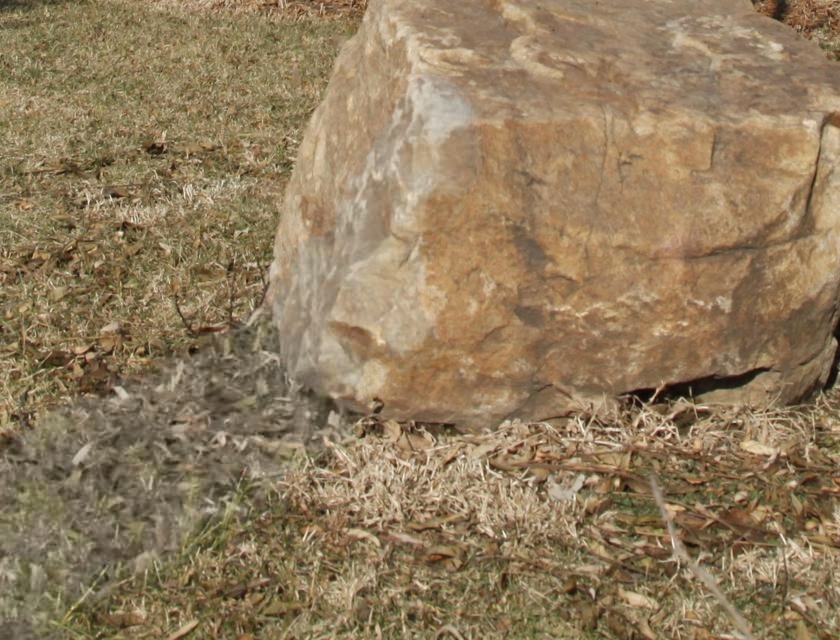}\\[0.05cm]
        \includegraphics[width=2cm,height=1.5cm]{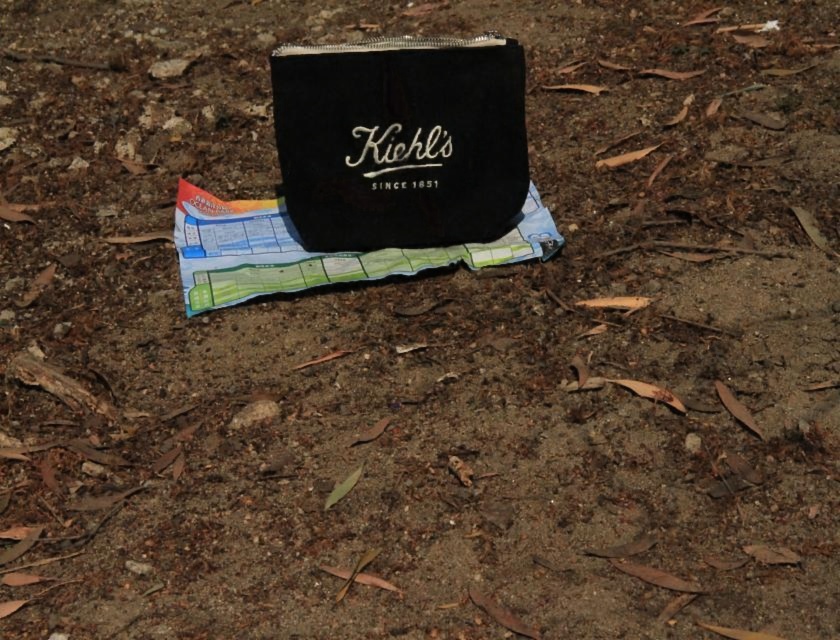}
    \end{subfigure}
    \begin{subfigure}[b]{0.115\linewidth}
        \centering
        \subcaption*{Inpaint4s \cite{li2023leveraging}}
        \includegraphics[width=2cm,height=1.5cm]{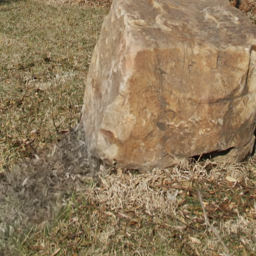}\\[0.05cm]
        \includegraphics[width=2cm,height=1.5cm]{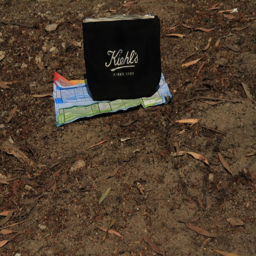}
    \end{subfigure}
    \begin{subfigure}[b]{0.115\linewidth}
        \centering
        \subcaption*{LFG-Diff \cite{mei2024latent}}
        \includegraphics[width=2cm,height=1.5cm]{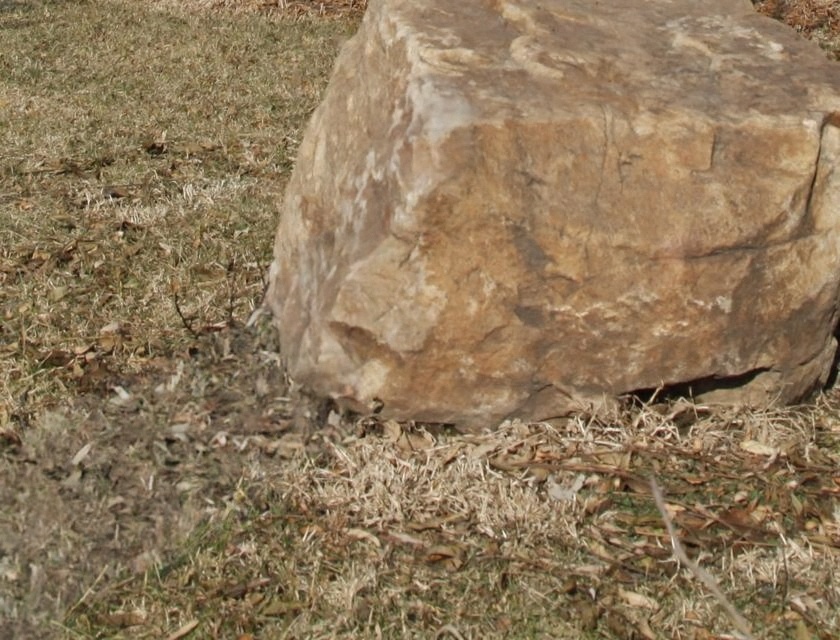}\\[0.05cm]
        \includegraphics[width=2cm,height=1.5cm]{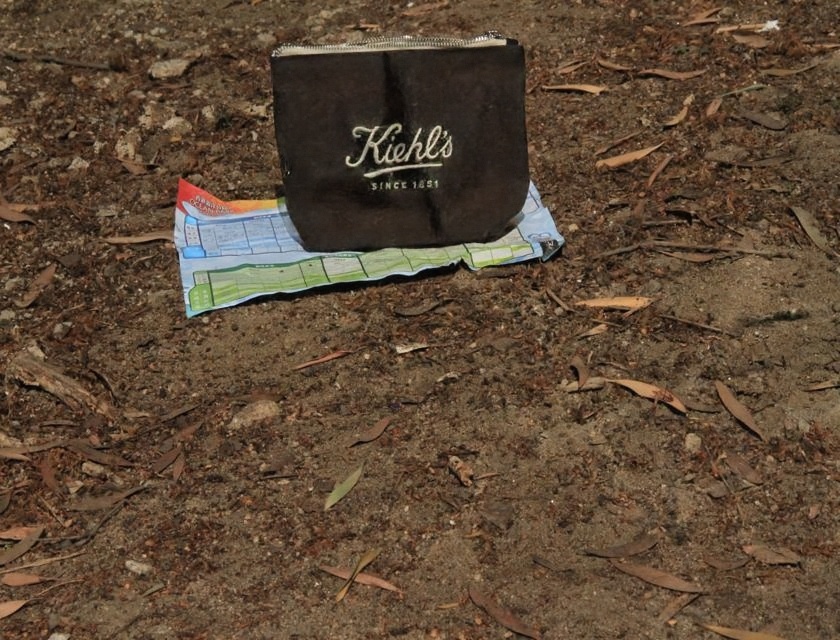}
    \end{subfigure}
    \begin{subfigure}[b]{0.115\linewidth}
        \centering
        \subcaption*{DeS3 \cite{jin2024des3}}
        \includegraphics[width=2cm,height=1.5cm]{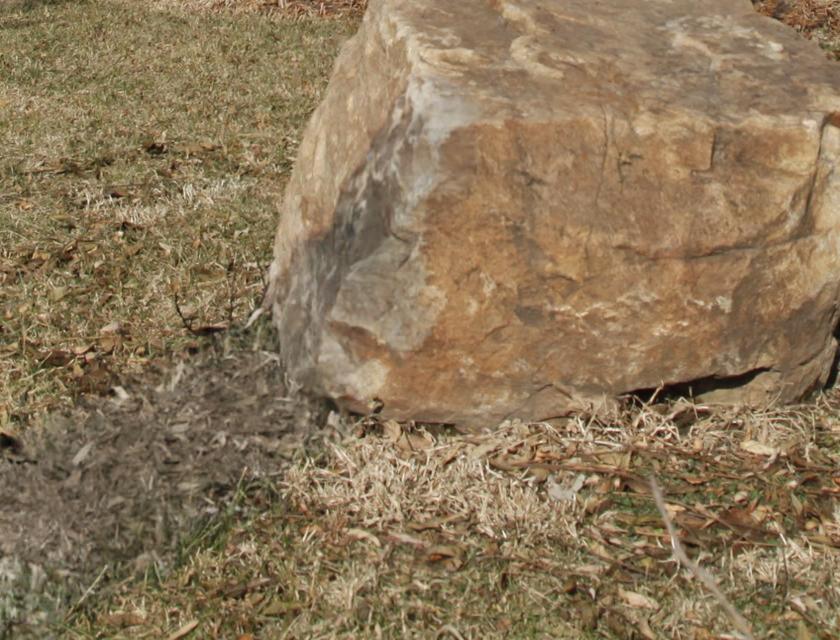}\\[0.05cm]
        \includegraphics[width=2cm,height=1.5cm]{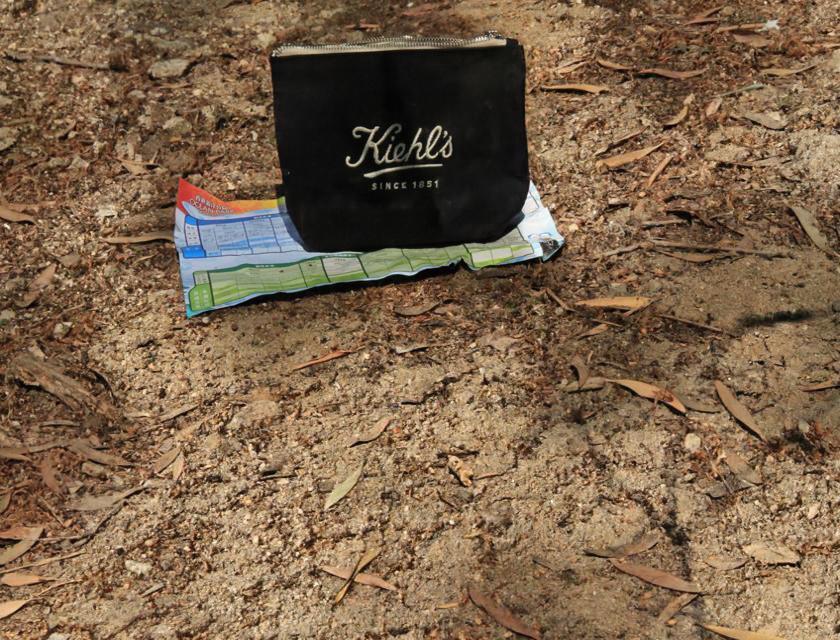}
    \end{subfigure}
    \begin{subfigure}[b]{0.115\linewidth}
        \centering
        \subcaption*{\textbf{Ours}}
        \includegraphics[width=2cm,height=1.5cm]{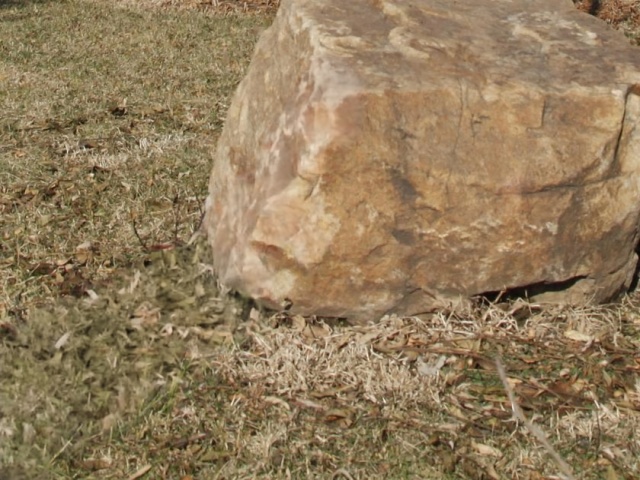}\\[0.05cm]
        \includegraphics[width=2cm,height=1.5cm]{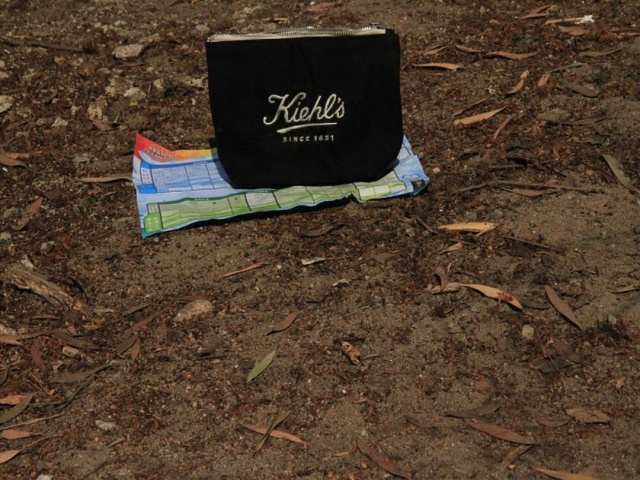}
    \end{subfigure}
    \begin{subfigure}[b]{0.115\linewidth}
        \centering
        \subcaption*{Ground Truth}
        \includegraphics[width=2cm,height=1.5cm]{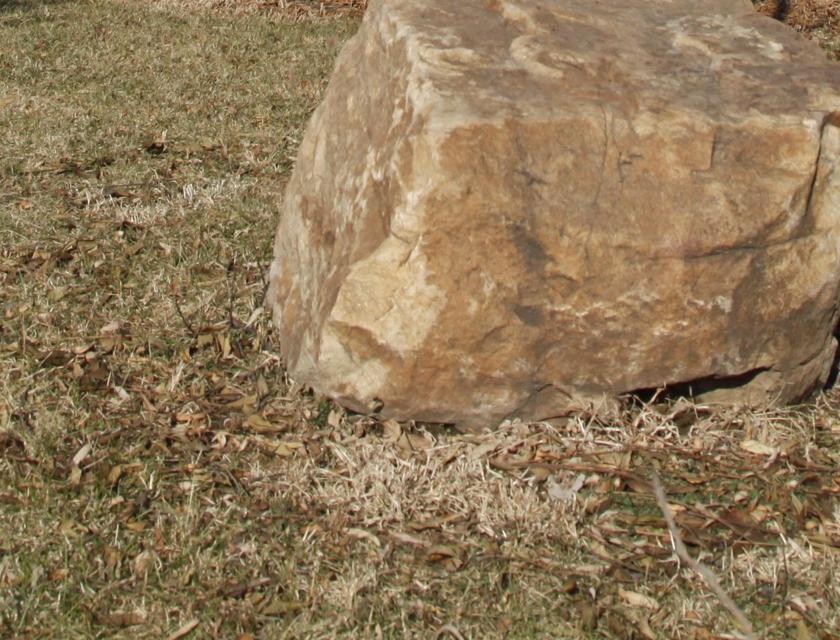}\\[0.05cm]
        \includegraphics[width=2cm,height=1.5cm]{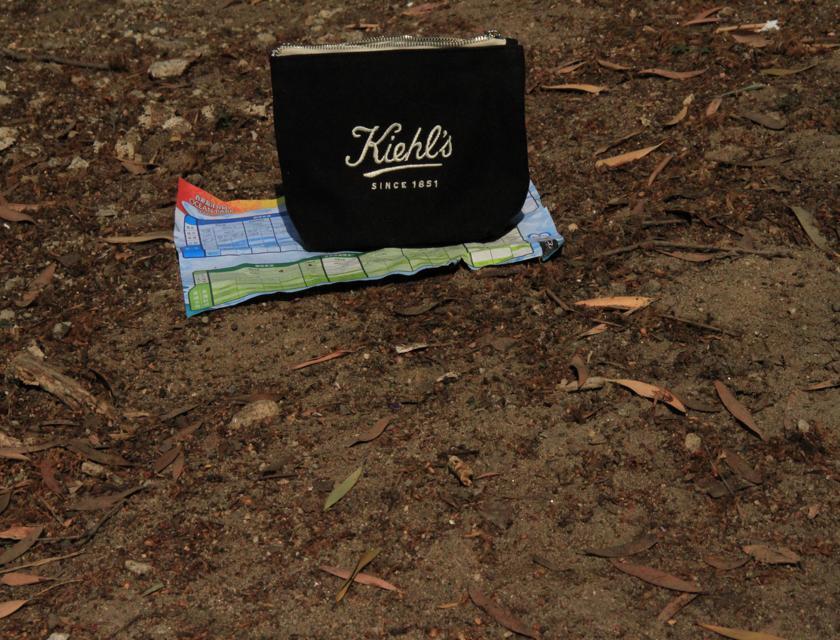}
    \end{subfigure}

    \caption{Qualitative results comparing ShadowMamba with other methods on the SRD dataset (model names abbreviated).} 
    \label{Fig 7}
\end{figure*}

To further demonstrate the advantages of ShadowMamba over competing methods, visual result comparisons on the AISTD and SRD datasets are shown in Figure \ref{Fig 6} and Figure \ref{Fig 7}. The results show that the proposed method produces fewer boundary artifacts and achieves a more natural balance between the restored shadow areas and the non-shadow areas.

\subsection{Ablation study}

To further analyze the structural design of ShadowMamba, the boundary-region selective scanning mechanism, and the mask denoising method, ablation studies are conducted on the AISTD \cite{le2019shadow} and SRD \cite{qu2017deshadownet} datasets.

\subsubsection{\textbf{Effectiveness of hierarchical combination}}

ShadowMamba adopts a hierarchical combination structure that extracts both local detail features and global brightness information, while avoiding additional computational cost and redundant information. Besides the default BRSSB-GSSB combination, this study also explores various block arrangements, with the experimental results shown in Table \ref{Tab5}. Meanwhile, the Dual-Branch State-Space Block (DSSB) is introduced, which fuses global and local scanning features in a single block and has been implemented in other studies \cite{zou2024freqmamba, wu2024rainmamba}. The results show that the hierarchical combination of BRSSB and GSSB (No.6) achieves the best performance. Compared with the DSSB structure (No.5), the hierarchical combination not only improves performance but also significantly reduces the number of parameters and computational complexity.

\begin{table*}[]

\caption{Ablation study for investigating different block arrangements of shadowMamba on the AISTD Dataset. (BR denotes BRSSB, G denotes GSSB, and D denotes DSSB. Each layer contains 2 blocks.)}
\centering

\label{Tab5}
\begin{tabular}{c|c|ccc}
\hline
\multirow{2}{*}{No.} & \multirow{2}{*}{Block Arrangement Combinations} & \multicolumn{1}{c}{ALL}  & \multicolumn{1}{c}{S}             & NS                                           \\ 
                &                                  & \multicolumn{1}{c}{PSNR$\uparrow$}  & \multicolumn{1}{c}{PSNR$\uparrow$}             & PSNR$\uparrow$            \\ \hline
 1                 & {[}G, G, G, G, G, G, G{]}              & \multicolumn{1}{c|}{35.21} & \multicolumn{1}{c|}{39.73}          & 38.80           \\
 2                 & {[}BR, G, G, G, G, G, BR{]}            & \multicolumn{1}{c|}{34.90} & \multicolumn{1}{c|}{39.53}          & 38.64          \\
 3                 & {[}BR, BR, BR, G, BR, BR, BR{]}        & \multicolumn{1}{c|}{34.92} & \multicolumn{1}{c|}{39.51}          & 38.66          \\
 4                 & {[}BR, BR, BR, BR, BR, BR, BR{]}       & \multicolumn{1}{c|}{35.13} & \multicolumn{1}{c|}{39.66}          & 38.74          \\
 5                 & {[}D, D, D, D, D, D, D{]}              & \multicolumn{1}{c|}{35.66} & \multicolumn{1}{c|}{39.90}          & 38.99           \\   \hline
\textbf{6}                 & \textbf{{[}BR, BR, G, G, G, BR, BR{]}} & \multicolumn{1}{c|}{\textbf{35.74}} & \multicolumn{1}{c|}{\textbf{39.96}} & \textbf{39.05} \\ \hline
\end{tabular}
\end{table*}

\subsubsection{\textbf{Effectiveness of each module}}

Table \ref{Tab6} presents the ablation results used to evaluate the effectiveness of each core module in ShadowMamba. The results show that removing any module leads to a drop in performance, and each module is both simple in structure and irreplaceable. This further demonstrates that BRSSM effectively extracts local details from shadow images, GSSM accurately models global brightness variations, and SFFN performs well in modeling the structure of the reordered sequences.

\begin{table}[]
\caption{Ablation study of different modules of ShadowMamba on the AISTD Dataset.}
\centering

\label{Tab6}
\begin{tabular}{c|c|ccc}
\hline
\multirow{2}{*}{No.}  & \multirow{2}{*}{Variant Models}  & \multicolumn{1}{c}{ALL}  & \multicolumn{1}{c}{S}             & NS                                           \\ 
                  &                                  & \multicolumn{1}{c}{PSNR$\uparrow$}  & \multicolumn{1}{c}{PSNR$\uparrow$}             & PSNR$\uparrow$           \\ \hline
1 & \multicolumn{1}{c|}{w/o BRSSM}                                 & \multicolumn{1}{c|}{35.01}          & \multicolumn{1}{c|}{39.36}          & 38.53\\
2 & \multicolumn{1}{c|}{w/o GSSM}                           & \multicolumn{1}{c|}{34.86}          & \multicolumn{1}{c|}{39.31}          & 38.44 \\
3 & \multicolumn{1}{c|}{w/o SFFN}                            & \multicolumn{1}{c|}{34.48}          & \multicolumn{1}{c|}{39.20}          & 38.35                              \\ \hline
\textbf{4} & \multicolumn{1}{c|}{\textbf{ShadowMamba}}        & \multicolumn{1}{c|}{\textbf{35.74}} & \multicolumn{1}{c|}{\textbf{39.96}} & \multicolumn{1}{c}{\textbf{39.05}} \\ \hline
\end{tabular}
\end{table}

\subsubsection{\textbf{Effectiveness of region-boundary scanning mechanism}}

To evaluate the effectiveness of the boundary-region selective scanning mechanism, this study replaces the scanning strategy in BRSSM while keeping the overall network structure unchanged. The experimental results are shown in Table \ref{Tab 6}. One variant, the region scanning mechanism, divides the image into two parts (shadow and non-shadow regions) and then applies global cross-scanning \cite{Liu2024VMambaVS}. Experimental comparisons show that both global scanning and region scanning fail to effectively model local details and boundary information, leading to weaker performance. Although local scanning \cite{huang2024localmamba} can capture fine details, it cannot make use of the semantic continuity between features. In contrast, the boundary-region selective scanning mechanism proposed in ShadowMamba not only extracts local detail features but also enhances semantic continuity among related pixels, thereby improving overall performance.

\begin{table*}[]
\centering

\caption{Ablation study of different scanning mechanisms on the AISTD dataset.}
\label{Tab 6}
\begin{tabular}{c|c|ccc}
\hline
\multirow{2}{*}{No.} & \multirow{2}{*}{Scanning Mechanism} & \multicolumn{1}{c}{ALL}  & \multicolumn{1}{c}{S}             & NS                                           \\ 
                  &                                  & \multicolumn{1}{c}{PSNR$\uparrow$}  & \multicolumn{1}{c}{PSNR$\uparrow$}             & PSNR$\uparrow$            \\ \hline
  1 & All Global scan                 & \multicolumn{1}{c|}{35.21}          & \multicolumn{1}{c|}{39.73}          & 38.80 \\
  2 & Region scan + Global scan                & \multicolumn{1}{c|}{35.23}          & \multicolumn{1}{c|}{39.75}          & 38.81 \\
  3 & Local scan + Global scan                 & \multicolumn{1}{c|}{35.46}          & \multicolumn{1}{c|}{39.81}          & 38.82
 \\
\hline
\textbf{4} & \multicolumn{1}{c|}{\textbf{Boundary-Region scan + Global scan}}           & \multicolumn{1}{c|}{\textbf{35.74}} & \multicolumn{1}{c|}{\textbf{39.96}} & \multicolumn{1}{c}{\textbf{39.05}} \\ \hline
\end{tabular}
\end{table*}

\subsubsection{\textbf{Effectiveness of mask denoising method}}

\begin{figure}[htbp]
    \centering
    \begin{subfigure}[b]{0.17\linewidth}
        \centering
        \subcaption*{L=0} 
        \includegraphics[width=1\textwidth,height=0.05\textheight]{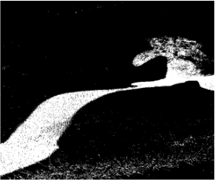}\\[0.05cm]
		\includegraphics[width=1\textwidth,height=0.05\textheight]{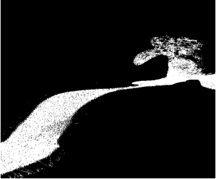}
    \end{subfigure}
    \begin{subfigure}[b]{0.17\linewidth}
        \centering
        \subcaption*{L=1}
        \includegraphics[width=1\textwidth,height=0.05\textheight]{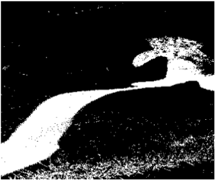}\\[0.05cm]
		\includegraphics[width=1\textwidth,height=0.05\textheight]{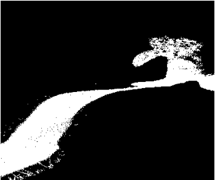}
    \end{subfigure}
    \begin{subfigure}[b]{0.17\linewidth}
        \centering
        \subcaption*{L=2}
        \includegraphics[width=1\textwidth,height=0.05\textheight]{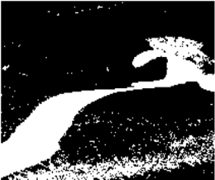}\\[0.05cm]
		\includegraphics[width=1\textwidth,height=0.05\textheight]{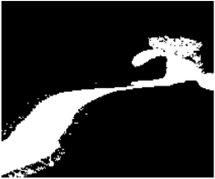}
    \end{subfigure}
    \begin{subfigure}[b]{0.17\linewidth}
        \centering
        \subcaption*{L=3}
        \includegraphics[width=1\textwidth,height=0.05\textheight]{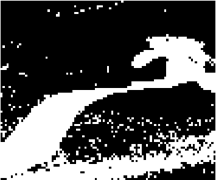}\\[0.05cm]
		\includegraphics[width=1\textwidth,height=0.05\textheight]{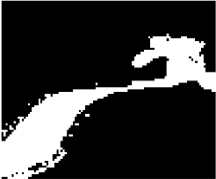}
    \end{subfigure}
    \begin{subfigure}[b]{0.17\linewidth}
        \centering
        \subcaption*{L=4}
        \includegraphics[width=1\textwidth,height=0.05\textheight]{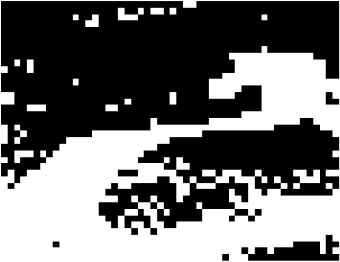}\\[0.05cm]
		\includegraphics[width=1\textwidth,height=0.05\textheight]{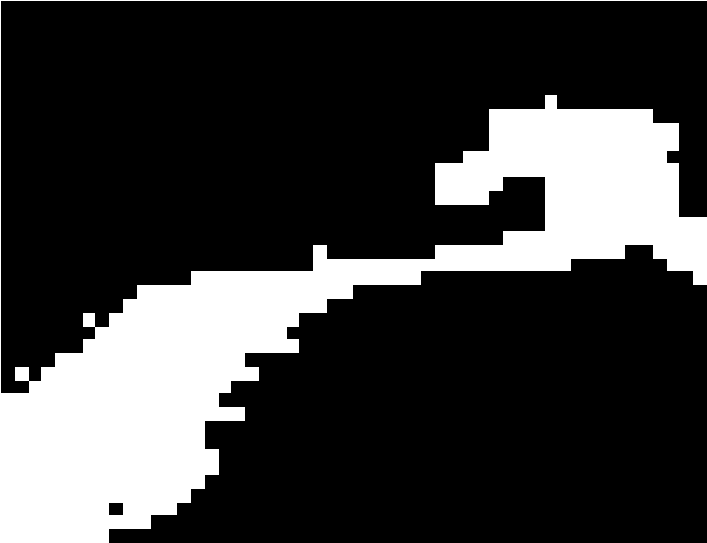}
    \end{subfigure}
    \caption{MaxPooling-based downsampling of the mask before (top) and after (bottom) denoising, where L denotes the number of downsampling layers.} 
     \label{Fig8}
\end{figure}

This study evaluates the effectiveness of the mask denoising method and the impact of noise using the SRD dataset \cite{qu2017deshadownet}, which contains heavy mask noise, as shown in Table \ref{Tab 8}. The results show that the proposed method can significantly improve the performance of models that rely on mask-guided interaction. ShadowFormer \cite{guo2023shadowformer} adopts a U-Net structure and introduces shadow interaction attention in its deeper layers (L2, L3, and L4). However, the MaxPooling operation used during downsampling amplifies noise in non-shadow regions, leading to poor performance on the SRD dataset. After applying the corrected mask, the model's performance improves significantly. Figure \ref{Fig8} compares the MaxPooling results before and after denoising, showing that heavy noise causes ShadowFormer's shadow interaction attention mechanism to fail.

\begin{table*}[]
\centering

\caption{Ablation study of mask denoising method on the SRD dataset.}
\label{Tab 8}
\begin{tabular}{c|c|ccc}
\hline
\multirow{2}{*}{No.}   & \multirow{2}{*}{Configuration} & \multicolumn{1}{c}{ALL}  & \multicolumn{1}{c}{S}             & NS                                           \\ 
                  &                                  & \multicolumn{1}{c}{PSNR$\uparrow$}  & \multicolumn{1}{c}{PSNR$\uparrow$}             & PSNR$\uparrow$            \\ \hline

  1 & ShadowFormer \cite{guo2023shadowformer} w/o mask denoising              & \multicolumn{1}{c|}{32.46}          & \multicolumn{1}{c|}{35.55}          & 36.82 \\ 
  2 & ShadowFormer w mask denoising             & \multicolumn{1}{c|}{34.13}          & \multicolumn{1}{c|}{38.76}          & 38.73 \\  

  3 & Local scan + Global scan w/o mask denoising              & \multicolumn{1}{c|}{34.84}          & \multicolumn{1}{c|}{36.78}          & 38,86 \\ 
  4 & Local scan + Global scan w mask denoising              & \multicolumn{1}{c|}{34.89}          & \multicolumn{1}{c|}{38.83}          & 38.90 \\ 
5 & ShadowMamba w/o mask denoising               & \multicolumn{1}{c|}{34.96}          & \multicolumn{1}{c|}{38.88}          & 38.97 \\\hline
  \textbf{6} & \textbf{ShadowMamba w mask denoising}           & \multicolumn{1}{c|}{\textbf{35.32}}          & \multicolumn{1}{c|}{\textbf{39.07}}          & \textbf{39.25} \\ \hline
\end{tabular}
\end{table*}

In addition, when the boundary-region selective scanning is replaced with local scanning \cite{huang2024localmamba}, the presence of noise has little impact on the local scanning variant. For ShadowMamba, its performance is slightly better than that of the local scanning variant even when noise is present. After mask denoising, ShadowMamba shows a clear improvement in performance. This suggests that even with noise, the boundary-region selective scanning mechanism can still retain the core function of local scanning, and its performance advantage becomes more evident as noise decreases.

\subsubsection{\textbf{Impact of window size}}

The choice of window size is crucial for the boundary-region selective scanning mechanism. Table \ref{Tab 9} shows the performance comparison under different window sizes. Experimental results indicate that when the window size is set to 8 or 10, the model achieves the best balance between accuracy and computational efficiency.

\begin{table}[]
\centering

\caption{Ablation study of window size on the AISTD dataset.}
\label{Tab 9}
\begin{tabular}{c|c|ccc}
\hline
\multirow{2}{*}{No.}   & \multirow{2}{*}{Window Size} & \multicolumn{1}{c}{ALL}  & \multicolumn{1}{c}{S}             & NS                                           \\ 
                  &                                  & \multicolumn{1}{c}{PSNR$\uparrow$}  & \multicolumn{1}{c}{PSNR$\uparrow$}             & PSNR$\uparrow$           \\ \hline
  1 & 4$\times$4                 & \multicolumn{1}{c|}{35.61}          & \multicolumn{1}{c|}{39.88}          & 38.92 \\
  2 & 8$\times$8               & \multicolumn{1}{c|}{\textbf{35.74}}          & \multicolumn{1}{c|}{\textbf{39.96}}          & 39.05 \\ 
  3 & 10$\times$10               & \multicolumn{1}{c|}{35.73}          & \multicolumn{1}{c|}{39.95}          & \textbf{39.07} \\ 
  4 & 16$\times$16                & \multicolumn{1}{c|}{35.65}          & \multicolumn{1}{c|}{39.82}          & 39.03 \\ \hline
\end{tabular}
\end{table}

\subsection{Discussion and future work}

The boundary-region selective scanning mechanism aims to shorten the distance between pixels of the same category in a sequence, thereby enhancing semantic continuity and improving shadow removal performance. The main idea is that information in sequence modeling is exchanged more easily between adjacent positions. Unlike Transformers, which use fully connected attention to directly establish relationships between tokens, this mechanism models the state of each token through state-space equations, and the scan operation allows information to be passed sequentially along the sequence. As a result, information spreads faster and has a stronger influence between nearby tokens, while distant tokens require longer transmission paths and experience greater information decay. Therefore, arranging semantically related pixels closer together in the sequence helps the model better capture semantic structures, improving its ability to model boundary and regional features, and ultimately enhancing shadow removal performance.

ShadowMamba has good scalability and can be applied to other mask-based tasks, such as image matting and watermark removal, to enhance the model's perception and understanding of local boundary information. In the future, it will be developed into a Mamba-based generative shadow removal framework, optimized for the characteristics of shadow images, to further improve shadow removal performance.

\section{Conclusion}

This paper presents a shadow removal model based on the Mamba architecture, called ShadowMamba, which is the first to apply Mamba to this task. The model adopts a hierarchical U-Net structure, with the shallow layers extracting local features and the deep layers capturing global brightness information, effectively reducing computational cost. Based on the characteristics of shadow images, a boundary-region selective scanning mechanism is designed to classify and rearrange the divided windows. This enhances the semantic continuity of pixels in similar regions and improves the model's ability to capture local details. In addition, a simple and effective mask denoising method is proposed, which not only improves the accuracy of the boundary-region selective scanning mechanism but also boosts the performance of other mask-guided models. Experimental results show that ShadowMamba outperforms existing methods in both hard and soft shadow scenarios, while also having advantages in parameter count and computational complexity.

\section*{Credit author statement}
Xiujin Zhu: Conceptualization, Data Curation, Investigation, Methodology, Visualization, Writing – Original draft. 
Chee-Onn Chow: Conceptualization, Validation, Supervision, Writing – Review \& Editing. 
Joon Huang Chuah: Supervision, Writing – Review \& Editing.

\section*{Declaration of competing interests}
The authors declare that they have no known competing financial interests or personal relationships that could have appeared to influence the work reported in this paper.

\section*{Data availability}
The processed data required to reproduce the findings of this study are available at \url{https://github.com/ZHUXIUJINChris/ShadowMamba}.

\bibliographystyle{cas-model2-names}

\bibliography{Reference}

\begin{thebibliography}{50}
\expandafter\ifx\csname natexlab\endcsname\relax\def\natexlab#1{#1}\fi
\providecommand{\url}[1]{\texttt{#1}}
\providecommand{\href}[2]{#2}
\providecommand{\path}[1]{#1}
\providecommand{\DOIprefix}{doi:}
\providecommand{\ArXivprefix}{arXiv:}
\providecommand{\URLprefix}{URL: }
\providecommand{\Pubmedprefix}{pmid:}
\providecommand{\doi}[1]{\href{http://dx.doi.org/#1}{\path{#1}}}
\providecommand{\Pubmed}[1]{\href{pmid:#1}{\path{#1}}}
\providecommand{\bibinfo}[2]{#2}
\ifx\xfnm\relax \def\xfnm[#1]{\unskip,\space#1}\fi
\bibitem[{Charbonnier et~al.(1994)Charbonnier, Blanc-Feraud, Aubert and
  Barlaud}]{charbonnier1994two}
\bibinfo{author}{Charbonnier, P.}, \bibinfo{author}{Blanc-Feraud, L.},
  \bibinfo{author}{Aubert, G.}, \bibinfo{author}{Barlaud, M.},
  \bibinfo{year}{1994}.
\newblock \bibinfo{title}{Two deterministic half-quadratic regularization
  algorithms for computed imaging}, in: \bibinfo{booktitle}{Proceedings of 1st
  international conference on image processing}, \bibinfo{organization}{IEEE}.
  pp. \bibinfo{pages}{168--172}.
\bibitem[{Chen et~al.(2023)Chen, Sun, Song and Luo}]{chen2023diffusiondet}
\bibinfo{author}{Chen, S.}, \bibinfo{author}{Sun, P.}, \bibinfo{author}{Song,
  Y.}, \bibinfo{author}{Luo, P.}, \bibinfo{year}{2023}.
\newblock \bibinfo{title}{Diffusiondet: Diffusion model for object detection},
  in: \bibinfo{booktitle}{Proceedings of the IEEE/CVF international conference
  on computer vision}, pp. \bibinfo{pages}{19830--19843}.
\bibitem[{Chen et~al.(2021)Chen, Long, Zhang and Xiao}]{chen2021canet}
\bibinfo{author}{Chen, Z.}, \bibinfo{author}{Long, C.}, \bibinfo{author}{Zhang,
  L.}, \bibinfo{author}{Xiao, C.}, \bibinfo{year}{2021}.
\newblock \bibinfo{title}{Canet: A context-aware network for shadow removal},
  in: \bibinfo{booktitle}{Proceedings of the IEEE/CVF international conference
  on computer vision}, pp. \bibinfo{pages}{4743--4752}.
\bibitem[{Cun et~al.(2020)Cun, Pun and Shi}]{cun2020towardsss}
\bibinfo{author}{Cun, X.}, \bibinfo{author}{Pun, C.M.}, \bibinfo{author}{Shi,
  C.}, \bibinfo{year}{2020}.
\newblock \bibinfo{title}{Towards ghost-free shadow removal via dual
  hierarchical aggregation network and shadow matting gan}, in:
  \bibinfo{booktitle}{Proceedings of the AAAI Conference on Artificial
  Intelligence}, pp. \bibinfo{pages}{10680--10687}.
\bibitem[{Einy et~al.(2022)Einy, Immer, Vered and Avidan}]{einy2022physics}
\bibinfo{author}{Einy, T.}, \bibinfo{author}{Immer, E.},
  \bibinfo{author}{Vered, G.}, \bibinfo{author}{Avidan, S.},
  \bibinfo{year}{2022}.
\newblock \bibinfo{title}{Physics based image deshadowing using local linear
  model}, in: \bibinfo{booktitle}{Proceedings of the IEEE/CVF conference on
  computer vision and pattern recognition}, pp. \bibinfo{pages}{3012--3020}.
\bibitem[{Gong and Cosker(2014)}]{gong2014interactive}
\bibinfo{author}{Gong, H.}, \bibinfo{author}{Cosker, D.}, \bibinfo{year}{2014}.
\newblock \bibinfo{title}{Interactive shadow removal and ground truth for
  variable scene categories}, in: \bibinfo{booktitle}{BMVC 2014-Proceedings of
  the British Machine Vision Conference 2014}.
\bibitem[{Gu and Dao(2023)}]{gu2023mamba}
\bibinfo{author}{Gu, A.}, \bibinfo{author}{Dao, T.}, \bibinfo{year}{2023}.
\newblock \bibinfo{title}{Mamba: Linear-time sequence modeling with selective
  state spaces}.
\newblock \bibinfo{journal}{arXiv preprint arXiv:2312.00752} .
\bibitem[{Gu et~al.(2020)Gu, Dao, Ermon, Rudra and R{\'e}}]{gu2020hippo}
\bibinfo{author}{Gu, A.}, \bibinfo{author}{Dao, T.}, \bibinfo{author}{Ermon,
  S.}, \bibinfo{author}{Rudra, A.}, \bibinfo{author}{R{\'e}, C.},
  \bibinfo{year}{2020}.
\newblock \bibinfo{title}{Hippo: Recurrent memory with optimal polynomial
  projections}.
\newblock \bibinfo{journal}{Advances in neural information processing systems}
  \bibinfo{volume}{33}, \bibinfo{pages}{1474--1487}.
\bibitem[{Gu et~al.(2021)Gu, Goel and R{\'e}}]{gu2021efficiently}
\bibinfo{author}{Gu, A.}, \bibinfo{author}{Goel, K.}, \bibinfo{author}{R{\'e},
  C.}, \bibinfo{year}{2021}.
\newblock \bibinfo{title}{Efficiently modeling long sequences with structured
  state spaces}.
\newblock \bibinfo{journal}{arXiv preprint arXiv:2111.00396} .
\bibitem[{Guo et~al.(2024)Guo, Li, Dai, Ouyang, Ren and Xia}]{guo2024mambair}
\bibinfo{author}{Guo, H.}, \bibinfo{author}{Li, J.}, \bibinfo{author}{Dai, T.},
  \bibinfo{author}{Ouyang, Z.}, \bibinfo{author}{Ren, X.},
  \bibinfo{author}{Xia, S.T.}, \bibinfo{year}{2024}.
\newblock \bibinfo{title}{Mambair: A simple baseline for image restoration with
  state-space model}, in: \bibinfo{booktitle}{European conference on computer
  vision}, \bibinfo{organization}{Springer}. pp. \bibinfo{pages}{222--241}.
\bibitem[{Guo et~al.(2023a)Guo, Huang, Liu, Cheng and
  Wen}]{guo2023shadowformer}
\bibinfo{author}{Guo, L.}, \bibinfo{author}{Huang, S.}, \bibinfo{author}{Liu,
  D.}, \bibinfo{author}{Cheng, H.}, \bibinfo{author}{Wen, B.},
  \bibinfo{year}{2023}a.
\newblock \bibinfo{title}{Shadowformer: Global context helps shadow removal},
  in: \bibinfo{booktitle}{Proceedings of the AAAI Conference on Artificial
  Intelligence}, pp. \bibinfo{pages}{710--718}.
\bibitem[{Guo et~al.(2023b)Guo, Wang, Yang, Huang, Wang, Pfister and
  Wen}]{guo2023shadowdiffusion}
\bibinfo{author}{Guo, L.}, \bibinfo{author}{Wang, C.}, \bibinfo{author}{Yang,
  W.}, \bibinfo{author}{Huang, S.}, \bibinfo{author}{Wang, Y.},
  \bibinfo{author}{Pfister, H.}, \bibinfo{author}{Wen, B.},
  \bibinfo{year}{2023}b.
\newblock \bibinfo{title}{Shadowdiffusion: When degradation prior meets
  diffusion model for shadow removal}, in: \bibinfo{booktitle}{Proceedings of
  the IEEE/CVF Conference on Computer Vision and Pattern Recognition}, pp.
  \bibinfo{pages}{14049--14058}.
\bibitem[{Guo et~al.(2023c)Guo, Wang, Yang, Wang and Wen}]{guo2023boundary}
\bibinfo{author}{Guo, L.}, \bibinfo{author}{Wang, C.}, \bibinfo{author}{Yang,
  W.}, \bibinfo{author}{Wang, Y.}, \bibinfo{author}{Wen, B.},
  \bibinfo{year}{2023}c.
\newblock \bibinfo{title}{Boundary-aware divide and conquer: A diffusion-based
  solution for unsupervised shadow removal}, in:
  \bibinfo{booktitle}{Proceedings of the IEEE/CVF International Conference on
  Computer Vision}, pp. \bibinfo{pages}{13045--13054}.
\bibitem[{Guo et~al.(2012)Guo, Dai and Hoiem}]{guo2012paired}
\bibinfo{author}{Guo, R.}, \bibinfo{author}{Dai, Q.}, \bibinfo{author}{Hoiem,
  D.}, \bibinfo{year}{2012}.
\newblock \bibinfo{title}{Paired regions for shadow detection and removal}.
\newblock \bibinfo{journal}{IEEE transactions on pattern analysis and machine
  intelligence} \bibinfo{volume}{35}, \bibinfo{pages}{2956--2967}.
\bibitem[{Hu et~al.(2024)Hu, Baumann, Gui, Grebenkova, Ma, Fischer and
  Ommer}]{hu2024zigma}
\bibinfo{author}{Hu, V.T.}, \bibinfo{author}{Baumann, S.A.},
  \bibinfo{author}{Gui, M.}, \bibinfo{author}{Grebenkova, O.},
  \bibinfo{author}{Ma, P.}, \bibinfo{author}{Fischer, J.},
  \bibinfo{author}{Ommer, B.}, \bibinfo{year}{2024}.
\newblock \bibinfo{title}{Zigma: A dit-style zigzag mamba diffusion model}, in:
  \bibinfo{booktitle}{European Conference on Computer Vision},
  \bibinfo{organization}{Springer}. pp. \bibinfo{pages}{148--166}.
\bibitem[{Huang et~al.(2024)Huang, Pei, You, Wang, Qian and
  Xu}]{huang2024localmamba}
\bibinfo{author}{Huang, T.}, \bibinfo{author}{Pei, X.}, \bibinfo{author}{You,
  S.}, \bibinfo{author}{Wang, F.}, \bibinfo{author}{Qian, C.},
  \bibinfo{author}{Xu, C.}, \bibinfo{year}{2024}.
\newblock \bibinfo{title}{Localmamba: Visual state space model with windowed
  selective scan}, in: \bibinfo{booktitle}{European Conference on Computer
  Vision}, \bibinfo{organization}{Springer}. pp. \bibinfo{pages}{12--22}.
\bibitem[{Jain et~al.(2023)Jain, Li, Chiu, Hassani, Orlov and
  Shi}]{jain2023oneformer}
\bibinfo{author}{Jain, J.}, \bibinfo{author}{Li, J.}, \bibinfo{author}{Chiu,
  M.T.}, \bibinfo{author}{Hassani, A.}, \bibinfo{author}{Orlov, N.},
  \bibinfo{author}{Shi, H.}, \bibinfo{year}{2023}.
\newblock \bibinfo{title}{Oneformer: One transformer to rule universal image
  segmentation}, in: \bibinfo{booktitle}{Proceedings of the IEEE/CVF Conference
  on Computer Vision and Pattern Recognition}, pp. \bibinfo{pages}{2989--2998}.
\bibitem[{Jin et~al.(2024)Jin, Ye, Yang, Yuan and Tan}]{jin2024des3}
\bibinfo{author}{Jin, Y.}, \bibinfo{author}{Ye, W.}, \bibinfo{author}{Yang,
  W.}, \bibinfo{author}{Yuan, Y.}, \bibinfo{author}{Tan, R.T.},
  \bibinfo{year}{2024}.
\newblock \bibinfo{title}{Des3: Adaptive attention-driven self and soft shadow
  removal using vit similarity}, in: \bibinfo{booktitle}{Proceedings of the
  AAAI Conference on Artificial Intelligence}, pp. \bibinfo{pages}{2634--2642}.
\bibitem[{Le and Samaras(2019)}]{le2019shadow}
\bibinfo{author}{Le, H.}, \bibinfo{author}{Samaras, D.}, \bibinfo{year}{2019}.
\newblock \bibinfo{title}{Shadow removal via shadow image decomposition}, in:
  \bibinfo{booktitle}{Proceedings of the IEEE/CVF International Conference on
  Computer Vision}, pp. \bibinfo{pages}{8578--8587}.
\bibitem[{Le and Samaras(2021)}]{le2021physics}
\bibinfo{author}{Le, H.}, \bibinfo{author}{Samaras, D.}, \bibinfo{year}{2021}.
\newblock \bibinfo{title}{Physics-based shadow image decomposition for shadow
  removal}.
\newblock \bibinfo{journal}{IEEE Transactions on Pattern Analysis and Machine
  Intelligence} \bibinfo{volume}{44}, \bibinfo{pages}{9088--9101}.
\bibitem[{Li et~al.(2024)Li, Li, Wang, He, Wang, Wang and
  Qiao}]{li2024videomamba}
\bibinfo{author}{Li, K.}, \bibinfo{author}{Li, X.}, \bibinfo{author}{Wang, Y.},
  \bibinfo{author}{He, Y.}, \bibinfo{author}{Wang, Y.}, \bibinfo{author}{Wang,
  L.}, \bibinfo{author}{Qiao, Y.}, \bibinfo{year}{2024}.
\newblock \bibinfo{title}{Videomamba: State space model for efficient video
  understanding}, in: \bibinfo{booktitle}{European conference on computer
  vision}, \bibinfo{organization}{Springer}. pp. \bibinfo{pages}{237--255}.
\bibitem[{Li et~al.(2023)Li, Guo, Abdelfattah, Lin, Feng, Tsang and
  Wang}]{li2023leveraging}
\bibinfo{author}{Li, X.}, \bibinfo{author}{Guo, Q.},
  \bibinfo{author}{Abdelfattah, R.}, \bibinfo{author}{Lin, D.},
  \bibinfo{author}{Feng, W.}, \bibinfo{author}{Tsang, I.},
  \bibinfo{author}{Wang, S.}, \bibinfo{year}{2023}.
\newblock \bibinfo{title}{Leveraging inpainting for single-image shadow
  removal}, in: \bibinfo{booktitle}{Proceedings of the IEEE/CVF International
  Conference on Computer Vision}, pp. \bibinfo{pages}{13055--13064}.
\bibitem[{Li et~al.(2025)Li, Xie, Jiang and Lu}]{li2025shadowmaskformer}
\bibinfo{author}{Li, Z.}, \bibinfo{author}{Xie, G.}, \bibinfo{author}{Jiang,
  G.}, \bibinfo{author}{Lu, Z.}, \bibinfo{year}{2025}.
\newblock \bibinfo{title}{Shadowmaskformer: Mask augmented patch embedding for
  shadow removal}.
\newblock \bibinfo{journal}{IEEE Transactions on Artificial Intelligence} .
\bibitem[{Liu et~al.(2023a)Liu, Wang, Fan, Li, Qu and Tang}]{liu2023decoupled}
\bibinfo{author}{Liu, J.}, \bibinfo{author}{Wang, Q.}, \bibinfo{author}{Fan,
  H.}, \bibinfo{author}{Li, W.}, \bibinfo{author}{Qu, L.},
  \bibinfo{author}{Tang, Y.}, \bibinfo{year}{2023}a.
\newblock \bibinfo{title}{A decoupled multi-task network for shadow removal}.
\newblock \bibinfo{journal}{IEEE Transactions on Multimedia}
  \bibinfo{volume}{25}, \bibinfo{pages}{9449--9463}.
\bibitem[{Liu et~al.(2023b)Liu, Guo, Fu, Ke, Xu, Feng, Tsang and
  Lau}]{liu2023structure}
\bibinfo{author}{Liu, Y.}, \bibinfo{author}{Guo, Q.}, \bibinfo{author}{Fu, L.},
  \bibinfo{author}{Ke, Z.}, \bibinfo{author}{Xu, K.}, \bibinfo{author}{Feng,
  W.}, \bibinfo{author}{Tsang, I.W.}, \bibinfo{author}{Lau, R.W.},
  \bibinfo{year}{2023}b.
\newblock \bibinfo{title}{Structure-informed shadow removal networks}.
\newblock \bibinfo{journal}{IEEE Transactions on Image Processing}
  \bibinfo{volume}{32}, \bibinfo{pages}{5823--5836}.
\bibitem[{Liu et~al.(2024a)Liu, Ke, Xu, Liu, Wang and Lau}]{liu2024recasting}
\bibinfo{author}{Liu, Y.}, \bibinfo{author}{Ke, Z.}, \bibinfo{author}{Xu, K.},
  \bibinfo{author}{Liu, F.}, \bibinfo{author}{Wang, Z.}, \bibinfo{author}{Lau,
  R.W.}, \bibinfo{year}{2024}a.
\newblock \bibinfo{title}{Recasting regional lighting for shadow removal}, in:
  \bibinfo{booktitle}{Proceedings of the AAAI Conference on Artificial
  Intelligence}, pp. \bibinfo{pages}{3810--3818}.
\bibitem[{Liu et~al.(2024b)Liu, Tian, Zhao, Yu, Xie, Wang, Ye, Jiao and
  Liu}]{Liu2024VMambaVS}
\bibinfo{author}{Liu, Y.}, \bibinfo{author}{Tian, Y.}, \bibinfo{author}{Zhao,
  Y.}, \bibinfo{author}{Yu, H.}, \bibinfo{author}{Xie, L.},
  \bibinfo{author}{Wang, Y.}, \bibinfo{author}{Ye, Q.}, \bibinfo{author}{Jiao,
  J.}, \bibinfo{author}{Liu, Y.}, \bibinfo{year}{2024}b.
\newblock \bibinfo{title}{Vmamba: Visual state space model}.
\newblock \bibinfo{journal}{Advances in neural information processing systems}
  \bibinfo{volume}{37}, \bibinfo{pages}{103031--103063}.
\bibitem[{Liu et~al.(2021)Liu, Lin, Cao, Hu, Wei, Zhang, Lin and
  Guo}]{liu2021swin}
\bibinfo{author}{Liu, Z.}, \bibinfo{author}{Lin, Y.}, \bibinfo{author}{Cao,
  Y.}, \bibinfo{author}{Hu, H.}, \bibinfo{author}{Wei, Y.},
  \bibinfo{author}{Zhang, Z.}, \bibinfo{author}{Lin, S.}, \bibinfo{author}{Guo,
  B.}, \bibinfo{year}{2021}.
\newblock \bibinfo{title}{Swin transformer: Hierarchical vision transformer
  using shifted windows}, in: \bibinfo{booktitle}{Proceedings of the IEEE/CVF
  international conference on computer vision}, pp.
  \bibinfo{pages}{10012--10022}.
\bibitem[{Mei et~al.(2024)Mei, Figueroa, Lin, Ding, Cohen and
  Patel}]{mei2024latent}
\bibinfo{author}{Mei, K.}, \bibinfo{author}{Figueroa, L.},
  \bibinfo{author}{Lin, Z.}, \bibinfo{author}{Ding, Z.},
  \bibinfo{author}{Cohen, S.}, \bibinfo{author}{Patel, V.M.},
  \bibinfo{year}{2024}.
\newblock \bibinfo{title}{Latent feature-guided diffusion models for shadow
  removal}, in: \bibinfo{booktitle}{Proceedings of the IEEE/CVF Winter
  Conference on Applications of Computer Vision}, pp.
  \bibinfo{pages}{4313--4322}.
\bibitem[{Niu et~al.(2022)Niu, Liu, Wu and Xing}]{niu2022boundary}
\bibinfo{author}{Niu, K.}, \bibinfo{author}{Liu, Y.}, \bibinfo{author}{Wu, E.},
  \bibinfo{author}{Xing, G.}, \bibinfo{year}{2022}.
\newblock \bibinfo{title}{A boundary-aware network for shadow removal}.
\newblock \bibinfo{journal}{IEEE Transactions on Multimedia}
  \bibinfo{volume}{25}, \bibinfo{pages}{6782--6793}.
\bibitem[{Qu et~al.(2017)Qu, Tian, He, Tang and Lau}]{qu2017deshadownet}
\bibinfo{author}{Qu, L.}, \bibinfo{author}{Tian, J.}, \bibinfo{author}{He, S.},
  \bibinfo{author}{Tang, Y.}, \bibinfo{author}{Lau, R.W.},
  \bibinfo{year}{2017}.
\newblock \bibinfo{title}{Deshadownet: A multi-context embedding deep network
  for shadow removal}, in: \bibinfo{booktitle}{Proceedings of the IEEE
  conference on computer vision and pattern recognition}, pp.
  \bibinfo{pages}{4067--4075}.
\bibitem[{Shi et~al.(2025)Shi, Xia, Jin, Wang, Zhao, Xia, Xiao and
  Yang}]{shi2025vmambair}
\bibinfo{author}{Shi, Y.}, \bibinfo{author}{Xia, B.}, \bibinfo{author}{Jin,
  X.}, \bibinfo{author}{Wang, X.}, \bibinfo{author}{Zhao, T.},
  \bibinfo{author}{Xia, X.}, \bibinfo{author}{Xiao, X.}, \bibinfo{author}{Yang,
  W.}, \bibinfo{year}{2025}.
\newblock \bibinfo{title}{Vmambair: Visual state space model for image
  restoration}.
\newblock \bibinfo{journal}{IEEE Transactions on Circuits and Systems for Video
  Technology} .
\bibitem[{Tao et~al.(2024)Tao, Liu, Dou, Muennighoff, Wan, Luo, Lin and
  Wong}]{tao2024scaling}
\bibinfo{author}{Tao, C.}, \bibinfo{author}{Liu, Q.}, \bibinfo{author}{Dou,
  L.}, \bibinfo{author}{Muennighoff, N.}, \bibinfo{author}{Wan, Z.},
  \bibinfo{author}{Luo, P.}, \bibinfo{author}{Lin, M.}, \bibinfo{author}{Wong,
  N.}, \bibinfo{year}{2024}.
\newblock \bibinfo{title}{Scaling laws with vocabulary: Larger models deserve
  larger vocabularies}.
\newblock \bibinfo{journal}{Advances in Neural Information Processing Systems}
  \bibinfo{volume}{37}, \bibinfo{pages}{114147--114179}.
\bibitem[{Touvron et~al.(2022)Touvron, Bojanowski, Caron, Cord, El-Nouby,
  Grave, Izacard, Joulin, Synnaeve, Verbeek et~al.}]{touvron2022resmlp}
\bibinfo{author}{Touvron, H.}, \bibinfo{author}{Bojanowski, P.},
  \bibinfo{author}{Caron, M.}, \bibinfo{author}{Cord, M.},
  \bibinfo{author}{El-Nouby, A.}, \bibinfo{author}{Grave, E.},
  \bibinfo{author}{Izacard, G.}, \bibinfo{author}{Joulin, A.},
  \bibinfo{author}{Synnaeve, G.}, \bibinfo{author}{Verbeek, J.}, et~al.,
  \bibinfo{year}{2022}.
\newblock \bibinfo{title}{Resmlp: Feedforward networks for image classification
  with data-efficient training}.
\newblock \bibinfo{journal}{IEEE transactions on pattern analysis and machine
  intelligence} \bibinfo{volume}{45}, \bibinfo{pages}{5314--5321}.
\bibitem[{Wan et~al.(2022)Wan, Yin, Wu, Wu, Liu and Wang}]{wan2022style}
\bibinfo{author}{Wan, J.}, \bibinfo{author}{Yin, H.}, \bibinfo{author}{Wu, Z.},
  \bibinfo{author}{Wu, X.}, \bibinfo{author}{Liu, Y.}, \bibinfo{author}{Wang,
  S.}, \bibinfo{year}{2022}.
\newblock \bibinfo{title}{Style-guided shadow removal}, in:
  \bibinfo{booktitle}{European Conference on Computer Vision},
  \bibinfo{organization}{Springer}. pp. \bibinfo{pages}{361--378}.
\bibitem[{Wan et~al.(2024)Wan, Yin, Wu, Wu, Liu and Wang}]{wan2024crformer}
\bibinfo{author}{Wan, J.}, \bibinfo{author}{Yin, H.}, \bibinfo{author}{Wu, Z.},
  \bibinfo{author}{Wu, X.}, \bibinfo{author}{Liu, Z.}, \bibinfo{author}{Wang,
  S.}, \bibinfo{year}{2024}.
\newblock \bibinfo{title}{Crformer: A cross-region transformer for shadow
  removal}.
\newblock \bibinfo{journal}{Image and Vision Computing} ,
  \bibinfo{pages}{105273}.
\bibitem[{Wang et~al.(2018)Wang, Li and Yang}]{wang2018stacked}
\bibinfo{author}{Wang, J.}, \bibinfo{author}{Li, X.}, \bibinfo{author}{Yang,
  J.}, \bibinfo{year}{2018}.
\newblock \bibinfo{title}{Stacked conditional generative adversarial networks
  for jointly learning shadow detection and shadow removal}, in:
  \bibinfo{booktitle}{Proceedings of the IEEE conference on computer vision and
  pattern recognition}, pp. \bibinfo{pages}{1788--1797}.
\bibitem[{Wu et~al.(2024)Wu, Yang, Xu, Wang, Zhou and Zhu}]{wu2024rainmamba}
\bibinfo{author}{Wu, H.}, \bibinfo{author}{Yang, Y.}, \bibinfo{author}{Xu, H.},
  \bibinfo{author}{Wang, W.}, \bibinfo{author}{Zhou, J.}, \bibinfo{author}{Zhu,
  L.}, \bibinfo{year}{2024}.
\newblock \bibinfo{title}{Rainmamba: Enhanced locality learning with state
  space models for video deraining}, in: \bibinfo{booktitle}{Proceedings of the
  32nd ACM International Conference on Multimedia}, pp.
  \bibinfo{pages}{7881--7890}.
\bibitem[{Xiao et~al.(2023)Xiao, Fu, Zhou, Liu and Zha}]{xiao2023random}
\bibinfo{author}{Xiao, J.}, \bibinfo{author}{Fu, X.}, \bibinfo{author}{Zhou,
  M.}, \bibinfo{author}{Liu, H.}, \bibinfo{author}{Zha, Z.J.},
  \bibinfo{year}{2023}.
\newblock \bibinfo{title}{Random shuffle transformer for image restoration},
  in: \bibinfo{booktitle}{International conference on machine learning},
  \bibinfo{organization}{PMLR}. pp. \bibinfo{pages}{38039--38058}.
\bibitem[{Xiao et~al.(2024)Xiao, Fu, Zhu, Li, Huang, Zhu and
  Zha}]{xiao2024homoformer}
\bibinfo{author}{Xiao, J.}, \bibinfo{author}{Fu, X.}, \bibinfo{author}{Zhu,
  Y.}, \bibinfo{author}{Li, D.}, \bibinfo{author}{Huang, J.},
  \bibinfo{author}{Zhu, K.}, \bibinfo{author}{Zha, Z.J.}, \bibinfo{year}{2024}.
\newblock \bibinfo{title}{Homoformer: Homogenized transformer for image shadow
  removal}, in: \bibinfo{booktitle}{Proceedings of the IEEE/CVF Conference on
  Computer Vision and Pattern Recognition}, pp. \bibinfo{pages}{25617--25626}.
\bibitem[{Xu et~al.(2025)Xu, Li, Zheng, Huang, Gu, Xu and Xu}]{xu2025omnisr}
\bibinfo{author}{Xu, J.}, \bibinfo{author}{Li, Z.}, \bibinfo{author}{Zheng,
  Y.}, \bibinfo{author}{Huang, C.}, \bibinfo{author}{Gu, R.},
  \bibinfo{author}{Xu, W.}, \bibinfo{author}{Xu, G.}, \bibinfo{year}{2025}.
\newblock \bibinfo{title}{Omnisr: Shadow removal under direct and indirect
  lighting}, in: \bibinfo{booktitle}{Proceedings of the AAAI Conference on
  Artificial Intelligence}, pp. \bibinfo{pages}{8887--8895}.
\bibitem[{Yang et~al.(2025)Yang, Xing, Yu, Fu, Huang and Zhu}]{yang2024vivim}
\bibinfo{author}{Yang, Y.}, \bibinfo{author}{Xing, Z.}, \bibinfo{author}{Yu,
  L.}, \bibinfo{author}{Fu, H.}, \bibinfo{author}{Huang, C.},
  \bibinfo{author}{Zhu, L.}, \bibinfo{year}{2025}.
\newblock \bibinfo{title}{Vivim: a video vision mamba for ultrasound video
  segmentation}.
\newblock \bibinfo{journal}{IEEE Transactions on Circuits and Systems for Video
  Technology} .
\bibitem[{Zhang et~al.(2018)Zhang, Cisse, Dauphin and
  Lopez-Paz}]{zhang2018mixup}
\bibinfo{author}{Zhang, H.}, \bibinfo{author}{Cisse, M.},
  \bibinfo{author}{Dauphin, Y.N.}, \bibinfo{author}{Lopez-Paz, D.},
  \bibinfo{year}{2018}.
\newblock \bibinfo{title}{Mixup: beyond empirical risk minimization}, in:
  \bibinfo{booktitle}{International Conference on Learning Representations}.
\bibitem[{Zhang et~al.(2015)Zhang, Zhang and Xiao}]{zhang2015shadow}
\bibinfo{author}{Zhang, L.}, \bibinfo{author}{Zhang, Q.},
  \bibinfo{author}{Xiao, C.}, \bibinfo{year}{2015}.
\newblock \bibinfo{title}{Shadow remover: Image shadow removal based on
  illumination recovering optimization}.
\newblock \bibinfo{journal}{IEEE Transactions on Image Processing}
  \bibinfo{volume}{24}, \bibinfo{pages}{4623--4636}.
\bibitem[{Zhang et~al.(2025)Zhang, Zhang, Liu, Xiao, Qian, Ahmed, Ambikairajah,
  Li and Epps}]{zhang2025mamba}
\bibinfo{author}{Zhang, X.}, \bibinfo{author}{Zhang, Q.}, \bibinfo{author}{Liu,
  H.}, \bibinfo{author}{Xiao, T.}, \bibinfo{author}{Qian, X.},
  \bibinfo{author}{Ahmed, B.}, \bibinfo{author}{Ambikairajah, E.},
  \bibinfo{author}{Li, H.}, \bibinfo{author}{Epps, J.}, \bibinfo{year}{2025}.
\newblock \bibinfo{title}{Mamba in speech: Towards an alternative to
  self-attention}.
\newblock \bibinfo{journal}{IEEE Transactions on Audio, Speech and Language
  Processing} .
\bibitem[{Zhao et~al.(2024)Zhao, Lv, Xu, Wei, Wang, Dang, Liu and
  Chen}]{zhao2024detrs}
\bibinfo{author}{Zhao, Y.}, \bibinfo{author}{Lv, W.}, \bibinfo{author}{Xu, S.},
  \bibinfo{author}{Wei, J.}, \bibinfo{author}{Wang, G.}, \bibinfo{author}{Dang,
  Q.}, \bibinfo{author}{Liu, Y.}, \bibinfo{author}{Chen, J.},
  \bibinfo{year}{2024}.
\newblock \bibinfo{title}{Detrs beat yolos on real-time object detection}, in:
  \bibinfo{booktitle}{Proceedings of the IEEE/CVF Conference on Computer Vision
  and Pattern Recognition}, pp. \bibinfo{pages}{16965--16974}.
\bibitem[{Zhu et~al.(2024)Zhu, Liao, Zhang, Wang, Liu and Wang}]{zhu2024vision}
\bibinfo{author}{Zhu, L.}, \bibinfo{author}{Liao, B.}, \bibinfo{author}{Zhang,
  Q.}, \bibinfo{author}{Wang, X.}, \bibinfo{author}{Liu, W.},
  \bibinfo{author}{Wang, X.}, \bibinfo{year}{2024}.
\newblock \bibinfo{title}{Vision mamba: efficient visual representation
  learning with bidirectional state space model}, in:
  \bibinfo{booktitle}{Proceedings of the 41st International Conference on
  Machine Learning}.
\bibitem[{Zhu et~al.(2022a)Zhu, Huang, Fu, Zhao, Sun and
  Zha}]{zhu2022bijective}
\bibinfo{author}{Zhu, Y.}, \bibinfo{author}{Huang, J.}, \bibinfo{author}{Fu,
  X.}, \bibinfo{author}{Zhao, F.}, \bibinfo{author}{Sun, Q.},
  \bibinfo{author}{Zha, Z.J.}, \bibinfo{year}{2022}a.
\newblock \bibinfo{title}{Bijective mapping network for shadow removal}, in:
  \bibinfo{booktitle}{Proceedings of the IEEE/CVF Conference on Computer Vision
  and Pattern Recognition}, pp. \bibinfo{pages}{5627--5636}.
\bibitem[{Zhu et~al.(2022b)Zhu, Xiao, Fang, Fu, Xiong and
  Zha}]{zhu2022efficient}
\bibinfo{author}{Zhu, Y.}, \bibinfo{author}{Xiao, Z.}, \bibinfo{author}{Fang,
  Y.}, \bibinfo{author}{Fu, X.}, \bibinfo{author}{Xiong, Z.},
  \bibinfo{author}{Zha, Z.J.}, \bibinfo{year}{2022}b.
\newblock \bibinfo{title}{Efficient model-driven network for shadow removal},
  in: \bibinfo{booktitle}{Proceedings of the AAAI conference on artificial
  intelligence}, pp. \bibinfo{pages}{3635--3643}.
\bibitem[{Zou et~al.(2024)Zou, Yu, Huang and Zhao}]{zou2024freqmamba}
\bibinfo{author}{Zou, Z.}, \bibinfo{author}{Yu, H.}, \bibinfo{author}{Huang,
  J.}, \bibinfo{author}{Zhao, F.}, \bibinfo{year}{2024}.
\newblock \bibinfo{title}{Freqmamba: Viewing mamba from a frequency perspective
  for image deraining}, in: \bibinfo{booktitle}{ACM Multimedia 2024}.

\end{thebibliography}

\end{document}